\RequirePackage{fix-cm}
\documentclass[twocolumn]{svjour3}          
\smartqed  
\usepackage{graphicx}
\usepackage{mathptmx}      
\usepackage{times}
\usepackage{epsfig}
\usepackage{amsmath}
\usepackage{amssymb}
\usepackage{url}            
\usepackage{booktabs}       
\usepackage{amsfonts}       
\usepackage{nicefrac}       
\usepackage{microtype}      
\usepackage{algorithm}
\usepackage{algpseudocode}
\usepackage{graphicx}
\usepackage{amsmath}
\usepackage{multirow}
\usepackage{color}
\usepackage{subfigure}
\usepackage[authoryear,sort&compress,]{natbib}
\newcommand{\eg}{\emph{e.g.}}
\newcommand{\etal}{\emph{et.al.}}
\newcommand{\ie}{\emph{i.e.}}
\newcommand{\etc}{\emph{etc}}
\newcommand{\wrt}{\emph{w.r.t. }}
\newcommand{\tabincell}[2]{\begin{tabular}{@{}#1@{}}#2\end{tabular}}
\journalname{IJCV}

\begin{document}

\title{GhostNets on Heterogeneous Devices via Cheap Operations
}


\author{Kai Han$^{1,2}$ \and Yunhe Wang$^2$ \and Chang Xu$^3$ \and Jianyuan Guo$^{2,3}$ \and Chunjing Xu$^2$ \and\\ Enhua Wu$^{1,4}$ \and Qi Tian$^2$
}


\institute{$^1$State Key Lab. of Computer Science, ISCAS. \& University of Chinese Academy of Sciences. \\
           $^2$Huawei Noah's Ark Lab.\\
           $^3$The University of Sydney.\\
           $^4$University of Macau.\\
           \email{\{hankai,weh\}@ios.ac.cn,\{yunhe.wang,tian.qi1\}@huawei.com}
}

\date{Received: date / Accepted: date}

\maketitle

\begin{abstract}
Deploying convolutional neural networks (CNNs) on mobile devices is difficult due to the limited memory and computation resources. We aim to design efficient neural networks for heterogeneous devices including CPU and GPU, by exploiting the redundancy in feature maps, which has rarely been investigated in neural architecture design.

For CPU-like devices, we propose a novel CPU-efficient Ghost (C-Ghost) module to generate more feature maps from cheap operations. Based on a set of intrinsic feature maps, we apply a series of linear transformations with cheap cost to generate many ghost feature maps that could fully reveal information underlying intrinsic features. The proposed C-Ghost module can be taken as a plug-and-play component to upgrade existing convolutional neural networks. C-Ghost bottlenecks are designed to stack C-Ghost modules, and then the lightweight C-GhostNet can be easily established.
We further consider the efficient networks for GPU devices. Without involving too many GPU-inefficient operations (\eg, depth-wise convolution) in a building stage, we propose to utilize the stage-wise feature redundancy to formulate GPU-efficient Ghost (G-Ghost) stage structure. The features in a stage are split into two parts where the first part is processed using the original block with fewer output channels for generating intrinsic features, and the other are generated using cheap operations by exploiting stage-wise redundancy.
Experiments conducted on benchmarks demonstrate the effectiveness of the proposed C-Ghost module and the G-Ghost stage. C-GhostNet and G-GhostNet can achieve the optimal trade-off of accuracy and latency for CPU and GPU, respectively. Code is available at \url{https://github.com/huawei-noah/CV-Backbones}.
\end{abstract}

\section{Introduction}
\label{intro}

Deep convolutional neural networks have shown excellent performance on various computer vision tasks, such as image recognition~\citep{alexnet,a3m}, object detection~\citep{fasterrcnn,retinanet}, and semantic segmentation~\citep{deeplab}. Traditional CNNs usually need a large number of parameters and floating point operations (FLOPs) to achieve a satisfactory accuracy, \eg, ResNet50~\citep{resnet} has about $25.6$M parameters and requires $4.1$B FLOPs to process an image of size $224\times224$. Thus, the recent trend of deep neural network design is to explore portable and efficient network architectures with acceptable performance for mobile devices (\eg, smart phones and self-driving cars).

Over the years, a series of methods have been proposed to investigate compact deep neural networks such as network pruning~\citep{deepcompression,thinet}, low-bit quantization~\citep{xnor,jacob2018quantization}, knowledge distillation~\citep{Distill,you2017learning}, \etc. Han~\etal~\citep{deepcompression} proposed to prune the unimportant weights in neural networks. \citep{l1-pruning} utilized $\ell_1$-norm regularization to prune filters for efficient CNNs. ~\citep{xnor} quantized the weights and the activations to 1-bit data for achieving large compression and speed-up ratios. \citep{Distill} introduced knowledge distillation for transferring knowledge from a larger model to a smaller model. However, performance of these methods are often upper bounded by pre-trained deep neural networks that have been taken as their baselines.

\begin{figure}[tp]
	\vspace{-0em}
	\small
	\centering
	\includegraphics[width=0.9\linewidth]{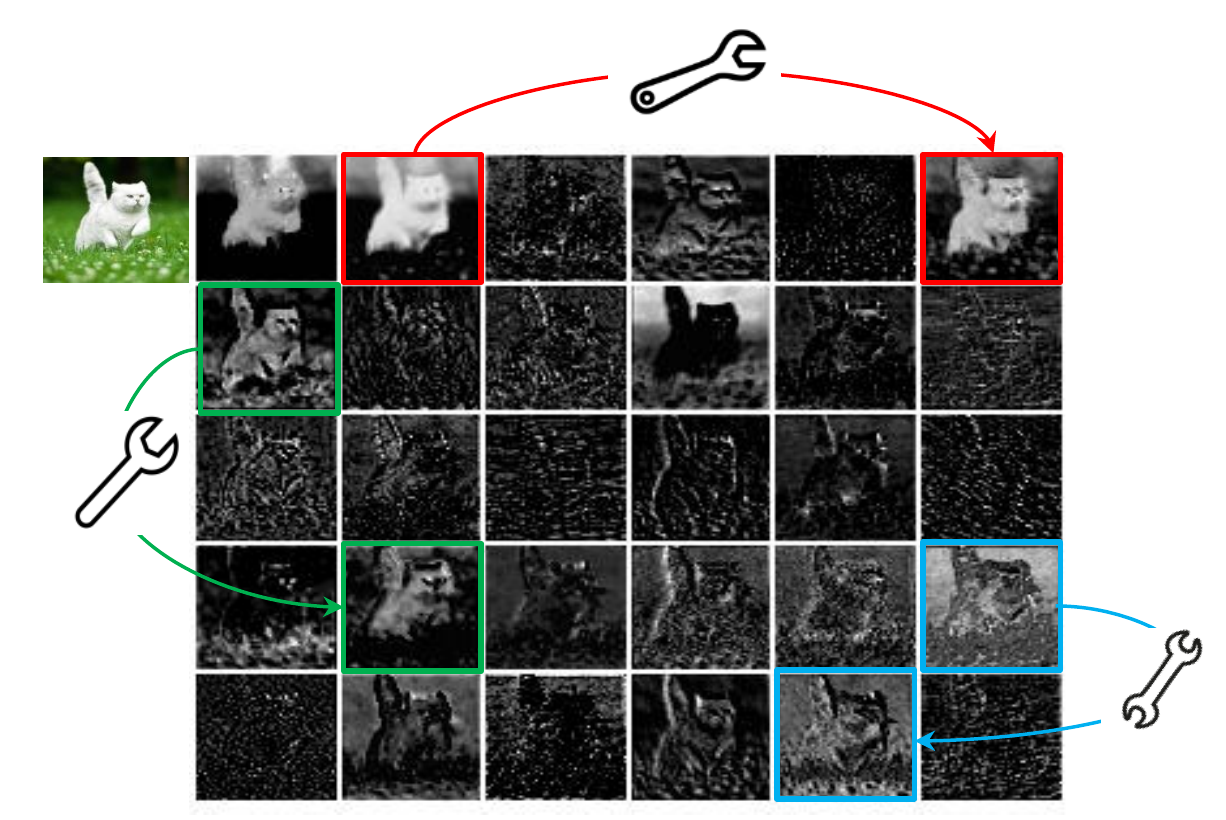}
	\caption{Visualization of some feature maps generated by the first residual group in ResNet50, where three similar feature map pair examples are annotated with boxes of the same color. One feature map in the pair can be approximately obtained by transforming the other one through cheap operations (denoted by spanners).}
	\label{Fig:maps}
	\vspace{-1em}
\end{figure}

Besides them, efficient neural architecture design has a very high potential for establishing highly efficient deep networks with fewer parameters and calculations, and recently has achieved considerable success. This kind of methods can also provide new search unit for automatic search methods~\citep{rlnas,cars,chen2019fasterseg}. For instance, MobileNet~\citep{mobilenet,mobilev2,mobilenetv3} utilized the depth-wise and pointwise convolutions to construct a unit for approximating the original convolutional layer with larger filters and achieved comparable performance. ShuffleNet~\citep{shufflenet,shufflev2} further explored a channel shuffle operation to enhance the performance of lightweight models.

Abundant and even redundant information in the feature maps of well-trained deep neural networks often guarantees a comprehensive understanding of the input data. For example, Figure~\ref{Fig:maps} presents some feature maps of an input image generated by ResNet50, and there exist many similar pairs of feature maps, like a \emph{ghost} of each another. Redundancy in feature maps could be an important characteristic for a successful deep neural network. Instead of avoiding the redundant feature maps, we tend to embrace them, but in a cost-efficient way. 

In this paper, we aim to design efficient neural networks for heterogeneous devices including CPU and GPU. For CPU devices, we introduce a novel CPU-efficient Ghost (C-Ghost) module to generate more features by using fewer parameters. Specifically, an ordinary convolutional layer in deep neural networks will be split into two parts. The first part involves ordinary convolutions but their total number will be rigorously controlled. Given the intrinsic feature maps from the first part, a series of simple linear operations are then applied for generating more feature maps. Without changing the size of output feature map, the overall required number of parameters and computational complexities in this C-Ghost module have been decreased, compared with those in vanilla convolutional neural networks. Based on C-Ghost module, we establish an efficient neural architecture, namely, C-GhostNet. We first replace original convolutional layers in benchmark neural architectures to demonstrate the effectiveness of C-Ghost modules, and then verify the superiority of our C-GhostNet on several benchmark visual datasets. Experimental results show that, the proposed C-Ghost module is able to decrease computational costs of generic convolutional layer while preserving similar recognition performance, and C-GhostNet can surpass state-of-the-art efficient deep models such as MobileNetV3~\citep{mobilenetv3}, on various tasks with fast inference on mobile devices.

Moreover, we consider designing efficient CNNs for GPU devices. Most of existing FLOPs-aware lightweight architectures including the aforementioned C-GhostNet are designed for CPUs. Because the significant differences on hardware architectures of CPU and GPU, some of operations with fewer FLOPs (especially the widely-used depth-wise convolution and channel shuffle~\citep{mobilenet,shufflenet}) are not perfectly efficient on GPU. In practice, these operators are usually of lower arithmetic intensity, \ie, ratio of computation to memory operations, which cannot be sufficiently utilized by the parallel computing capability~\citep{roofline,squeezenext}. Thus, the reduction on latency of using them on GPUs is always not up to our expectations. 
In this paper, to improve the latency performance of CNN architectures on GPU devices, we present a new GPU-efficient Ghost (G-Ghost) stage for building efficient networks. Instead of decomposing each convolutional layer to several sub-layers with fewer FLOPs and parameters, we investigate the stage-wise feature redundancy across blocks in the conventional structure. Basically, a building stage in modern CNN architectures usually consists of several convolutional layers or blocks. Since intermediate features in an arbitrary building stage are of the same size (\eg, 28$\times$28 in the second stage of ResNet), the feature similarity and redundancy not only exist inside one layer but also exist across multiple layers in the stage (Figure~\ref{Fig:feature-maps}). Therefore, we propose to generate intrinsic features utilizing the original stage with fewer number of channels and use cheap operation on GPUs to directly produce ghost features. To enhance the feature representation ability, we maximally excavate the information in intermediate layers and additively aggregate it into ghost features. The new G-Ghost stage can be well embedded into any mainstream CNN architectures. For the vanilla C-GhostNet, we eliminate all C-Ghost modules and using the proposed G-Ghost stage to replace original ones for establishing lightweight G-GhostNet. Experiments conducted on several benchmark models and datasets demonstrate the effectiveness of the proposed G-Ghost stage. The G-GhostNet architecture obtains highest inference speed on GPUs with state-of-the-art performance for image recognition.

The preliminary version of this work was presented as~\citep{ghostnet} on CVPR 2020. In this journal version, we extend the original work in significant ways. First, to promote the CNN speed on GPUs, we further introduce the G-Ghost stage structure by exploiting stage-wise feature redundancy. A mix operation is proposed for better preserving the intermediate representation inside a stage. Second, we build the  GPU-efficient G-GhostNet architecture with G-Ghost stages, which provides a lightweight CNN with state-of-the-art performance on GPUs. Third, extensive experiments on different networks and visual tasks are conducted to demonstrate the effectiveness of our models.

\section{Related Work}\label{related-work}
Here we revisit the existing methods for lightening neural networks in two parts: model compression and compact model design.

\textbf{
\subsection{Model Compression}
}
For a given neural network, model compression aims to reduce the computation, energy and storage cost~\citep{deepcompression,wang2019e2,gui2019model,xu2019positive,addernet}. Pruning connections~\citep{han2015learning,deepcompression,cnnpack} cuts out the unimportant connections between neurons. Channel pruning~\citep{wen2016learning,cp,l1-pruning,thinet,nisp,huang2018data,liu2019learning} further targets on removing useless channels for easier acceleration in practice. Model quantization~\citep{xnor,bnn,jacob2018quantization,yang2020searching} represents weights or activations in neural networks with discrete values for compression and calculation acceleration. Specifically, binarization methods~\citep{bnn,xnor,bireal,shen2019searching,han2020training} with only 1-bit values can extremely accelerate the model by efficient binary operations. Tensor decomposition~\citep{jaderberg2014speeding,denton2014exploiting,han2021learning} reduces the parameters or computation by exploiting the redundancy and low-rank property in weights. Knowledge distillation~\citep{Distill,han2018co,chen2019data,defeat} utilizes larger models to teach smaller ones, which improves the performance of smaller models. The performances of these methods usually depend on the given pre-trained models. The improvement on basic operations and architectures will make them go further.

\textbf{
\subsection{Compact Model Design}
}
With the need for deploying neural networks on embedded devices, a series of compact models are proposed in recent years~\citep{xception,mobilenet,mobilev2,mobilenetv3,shufflenet,shufflev2,shift,yang2019legonet}. Xception~\citep{xception} utilizes depth-wise convolution operation for more efficient use of model parameters. MobileNets~\citep{mobilenet} are a series of light weight deep neural networks based on depth-wise separable convolutions. MobileNetV2~\citep{mobilev2} proposes inverted residual block and MobileNetV3~\citep{mobilenetv3} further utilizes AutoML technology~\citep{rlnas,cars,gong2019autogan} achieving better performance with fewer FLOPs. ShuffleNet~\citep{shufflenet} introduces channel shuffle operation to improve the information flow exchange between channel groups. ShuffleNetV2~\citep{shufflev2} further considers the actual speed on target hardware for compact model design. Although these models obtain great performance with very few FLOPs, the correlation and redundancy between feature maps has never been well exploited.

In addition, there are also several works studying the inference efficiency of CNNs on GPUs. SqueezeNext~\citep{squeezenext} uses decomposed convolutions to reduce the complexity of residual block which is a kind of tensor decomposition technology. RegNet~\citep{regnet} observes that the GPU speed is much related to the activations and gives the empirically guidance for GPU-efficient models. In fact, the structure of RegNet basically follows that of ResNet. Instead, this paper proposes a new structure which can be used to improve most of the existing CNNs including RegNet.

\vspace{1em}
\section{CPU-Efficient GhostNet}\label{Approach}
In this section, we will first introduce the C-Ghost module to utilize a few small filters to generate more feature maps from the original convolutional layer, and then develop a new C-GhostNet with an extremely efficient architecture and high performance.

\textbf{
\subsection{Ghost Module for More Features}
}
Deep convolutional neural networks~\citep{alexnet,vgg,resnet} often consist of a large number of convolutions that result in massive computational costs. Although recent works such as MobileNet~\citep{mobilenet,mobilev2} and ShuffleNet~\citep{shufflev2} have introduced depth-wise convolution or shuffle operation to build efficient CNNs using smaller convolution filters (floating-number operations), the remaining $1\times1$ convolution layers would still occupy considerable memory and FLOPs.

Given the widely existing redundancy in intermediate feature maps calculated by mainstream CNNs as shown in Figure~\ref{Fig:maps}, we propose to reduce the required resources, \ie, convolution filters used for generating them. In practice, given the input data $X\in\mathbb{R}^{c\times h\times w}$, where $c$ is the number of input channels and $h$ and $w$ are the height and width of the input data, respectively,  the operation of an arbitrary convolutional layer for producing $n$ feature maps can be formulated as
\begin{equation}\label{eq:conv}
	Y = X*f+b,
\end{equation}
where $*$ is the convolution operation, $b$ is the bias term, $Y\in\mathbb{R}^{h'\times w'\times n}$ is the output feature map with $n$ channels, and $f\in\mathbb{R}^{c\times k\times k \times n}$ is the convolution filters in this layer. In addition, $h'$ and $w'$ are the height and width of the output data, and $k\times k$ is the kernel size of convolution filters $f$, respectively. During this convolution procedure, the required number of FLOPs can be calculated as $n\cdot h'\cdot w'\cdot c\cdot k\cdot k$, which is often as large as hundreds of thousands since the number of filters $n$ and the channel number $c$ are generally very large (\eg, 256 or 512).


\begin{figure}[ht]
	\vspace{-0.5em}
	\centering 
	\subfigure[The convolutional layer.] {
		\includegraphics[width=0.5\linewidth]{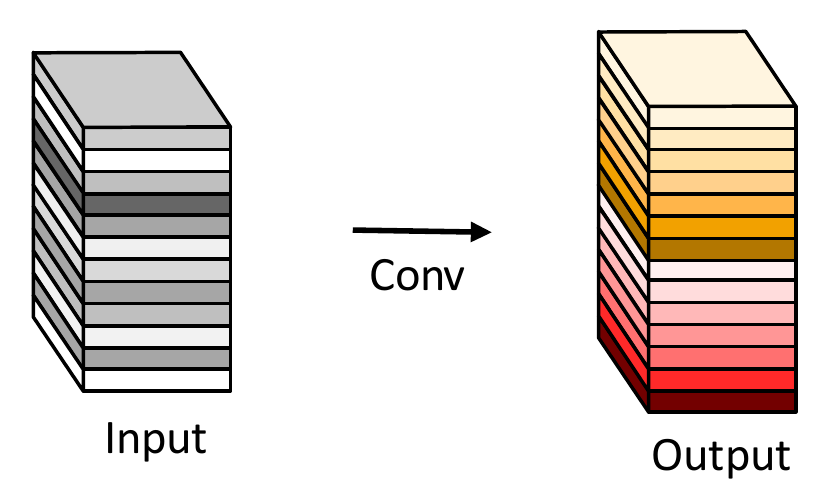}
	}
	\subfigure[The C-Ghost module.] {
		\includegraphics[width=0.9\linewidth]{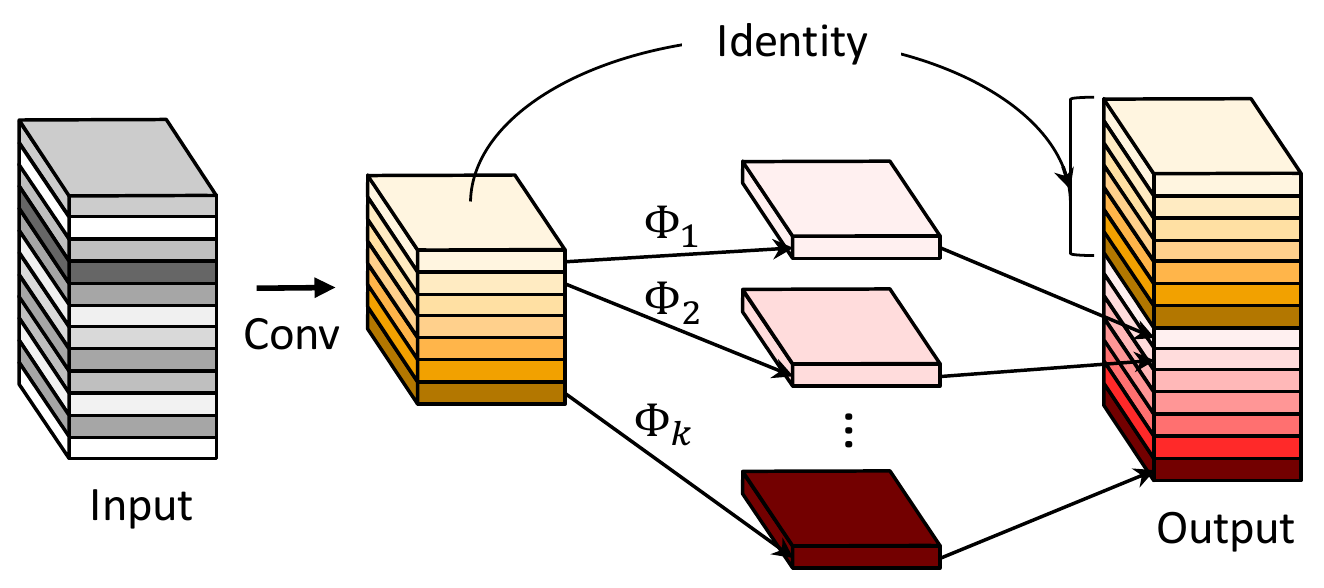} 
	}
	\caption{An illustration of the convolutional layer and the proposed C-Ghost module for outputting the same number of feature maps. $\Phi$ represents the cheap operation.} 
	\label{Fig:Ghost}
	\vspace{-0.5em}
\end{figure}

According to Eq.~\ref{eq:conv}, the number of parameters (in $f$ and $b$) to be optimized is explicitly determined by the dimensions of input and output feature maps. As observed in Figure~\ref{Fig:maps}, the output feature maps of convolutional layers often contain much redundancy, and some of them could be similar with each other. We point out that it is unnecessary to generate these redundant feature maps one by one with large number of FLOPs and parameters. Suppose that the output feature maps are ``ghosts'' of a handful of intrinsic feature maps generated by some cheap operations. These intrinsic feature maps are often of smaller size and produced by ordinary convolution filters. Specifically, $m$ intrinsic feature maps $Y'\in\mathbb{R}^{h'\times w'\times m}$ are generated using a primary convolution:
\begin{equation}
	Y' = X*f',
\end{equation}
where $f'\in\mathbb{R}^{c\times k\times k \times m}$ is the utilized filters, $m\leq n$ and the bias term is omitted for simplicity. The hyper-parameters such as filter size, stride, padding, are the same as those in the ordinary convolution (Eq.~\ref{eq:conv}) to keep the spatial size (\ie, $h'$ and $w'$) of the output feature maps consistent. To further obtain the desired $n$ feature maps, we propose to apply a series of cheap operations on each intrinsic feature in $Y'$ to generate $s$ ghost features according to the following function:
\begin{equation}
	y_{ij} = \Phi_{i,j}(y'_i),\quad \forall\; i = 1,...,m,\;\; j = 1,...,s,
	\label{eq:ghost}
\end{equation}
where $y'_i$ is the $i$-th intrinsic feature map in $Y'$, $\Phi_{i,j}$ in the above function is the $j$-th (except the last one) cheap operation for generating the $j$-th ghost feature map $y_{ij}$. Here $y'_i$ can have one or more ghost feature maps $\{y_{ij}\}_{j=1}^{s}$. The cheap operation $\Phi_{i,j}$ can perform local feature aggregation and adjusting to generate ghost feature maps. The last $\Phi_{i,s}$ is the identity mapping for preserving the intrinsic feature maps as shown in Figure~\ref{Fig:Ghost}(b). By utilizing Eq.~\ref{eq:ghost}, we can obtain $n=m\cdot s$ feature maps $Y=[y_{11},y_{12},\cdots,y_{ms}]$ as the output data of a C-Ghost module as shown in Figure~\ref{Fig:Ghost}(b). Note that the cheap operations $\Phi$ operate on each channel whose computational cost is much less than the ordinary convolution. In practice, there could be several different cheap operations in a C-Ghost module, \eg, $3\times 3$ and $5\times5$ linear kernels, which will be analyzed in the experiment part.

\paragraph{Difference from Existing  Methods.} The proposed C-Ghost module has major differences from existing efficient convolution schemes. i) Compared with the units in~\citep{mobilenet,shufflenet} which utilize $1\times 1$ pointwise convolution widely, the primary convolution in C-Ghost module can have customized kernel size. ii) Existing methods~\citep{mobilenet,mobilev2,shufflenet,shufflev2} adopt pointwise convolutions to process features across channels and then take depth-wise convolution to process spatial information. In contrast, C-Ghost module adopts ordinary convolution to first generate a few intrinsic feature maps, and then utilizes cheap linear operations to augment the features and increase the channels. iii)  The operation to process each feature map is limited to depth-wise convolution or shift operation in previous efficient architectures~\citep{mobilenet,shufflenet,shift,active-shift}, while linear operations in C-Ghost module can have large diversity. iv) In addition, the identity mapping is paralleled with linear transformations in C-Ghost module to preserve the intrinsic feature maps.

\paragraph{Analysis on Complexities.} Since we can utilize the proposed C-Ghost module in Eq.~\ref{eq:ghost} to generate the same number of feature maps as that of an ordinary convolutional layer, we can easily integrate the C-Ghost module into existing well designed neural architectures to reduce the computation costs. Here we further analyze the profit on memory usage and theoretical speed-up by employing the C-Ghost module. For example, there are $1$ identity mapping and $m \cdot (s-1) = \frac{n}{s}\cdot (s-1)$ linear operations, and the averaged kernel size of each linear operation is equal to $d\times d$. Ideally, the $n\cdot(s-1)$ linear operations can have different shapes and parameters, but the online inference will be obstructed especially considering the utility of hardware. Thus, we suggest to take linear operations of the same size (\eg, $3\times3$ or $5\times5$) in one C-Ghost module for efficient implementation. The theoretical speed-up ratio of upgrading ordinary convolution with the C-Ghost module is
\begin{equation}
	\begin{aligned}
		r_s &= \frac{n\cdot h'\cdot w'\cdot c\cdot k\cdot k}{\frac{n}{s}\cdot h'\cdot w'\cdot c\cdot k\cdot k + (s-1)\cdot \frac{n}{s}\cdot h'\cdot w'\cdot d\cdot d}\\
		&= \frac{c\cdot k\cdot k}{\frac{1}{s}\cdot c\cdot k\cdot k+\frac{s-1}{s}\cdot d\cdot d} \approx \frac{s\cdot c}{s+c-1}\approx s,
	\end{aligned}
	\label{eq:rs}
\end{equation}
where $d\times d$ has the similar magnitude as that of $k\times k$, and $s\ll c$. Similarly, the compression ratio can be calculated as
\begin{equation}
	\begin{aligned}
		r_c &= \frac{n\cdot c\cdot k\cdot k}{\frac{n}{s}\cdot c\cdot k\cdot k + (s-1)\cdot\frac{n}{s}\cdot d\cdot d} \approx \frac{s\cdot c}{s+c-1} \approx s,
	\end{aligned}
	\label{eq:rc}
\end{equation}
which is equal to that of the speed-up ratio by utilizing the proposed C-Ghost module.

\begin{figure}[htb]
	\vspace{-0em}
	\centering
	\includegraphics[width=0.78\linewidth]{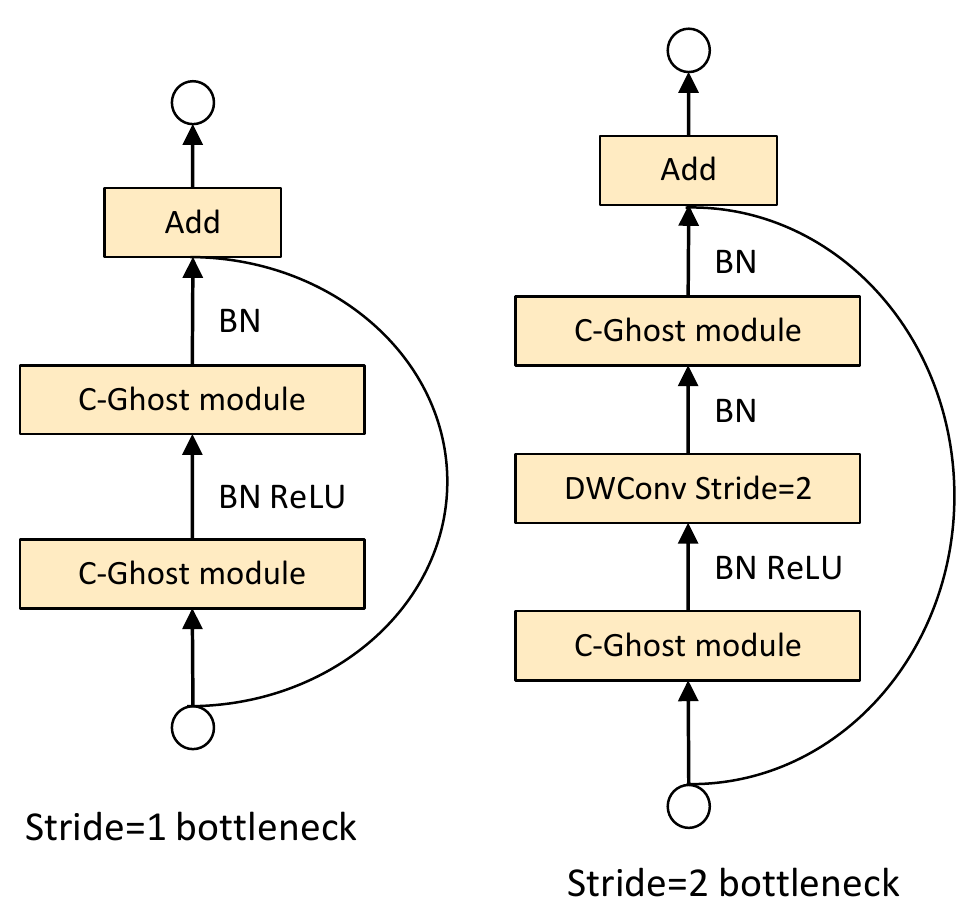}
	\vspace{-0.0em}
	\caption{C-Ghost bottleneck. Left: C-Ghost bottleneck with stride=1; right: C-Ghost bottleneck with stride=2.}
	\label{fig:hw}
	\vspace{-0em}
\end{figure}

\textbf{
\subsection{Building Lightweight C-GhostNet}
}
\paragraph{C-Ghost Bottlenecks.}
Taking the advantages of C-Ghost module, we introduce the C-Ghost bottleneck (G-bneck) specially designed for small CNNs. As shown in Figure~\ref{fig:hw}, the C-Ghost bottleneck appears to be similar to the basic residual block in ResNet~\citep{resnet} in which several convolutional layers and shortcuts are integrated. The proposed ghost bottleneck mainly consists of two stacked C-Ghost modules. The first C-Ghost module acts as an expansion layer increasing the number of channels. We refer the ratio between the number of the output channels and that of the input as \emph{expansion ratio}. The second C-Ghost module reduces the number of channels to match the shortcut path. Then the shortcut is connected between the inputs and the outputs of these two C-Ghost modules. The batch normalization (BN)~\citep{bn} and ReLU nonlinearity are applied after each layer, except that ReLU is not used after the second C-Ghost module as suggested by MobileNetV2~\citep{mobilev2}. The C-Ghost bottleneck described above is for stride=1. As for the case where stride=2, the shortcut path is implemented by a downsampling layer and a depth-wise convolution with stride=2 is inserted between the two C-Ghost modules. In practice, the primary convolution in C-Ghost module here is pointwise convolution for its efficiency.

\begin{table}[htb]
	\vspace{-0.1em}
	\centering
	\renewcommand\arraystretch{1.0}
	\caption{Overall architecture of C-GhostNet. G-bneck denotes C-Ghost bottleneck. \#exp means expansion size. \#out means the number of output channels. SE denotes whether using SE module.}
	\label{tab:C-GhostNet}
	\small
	\begin{tabular}{c|c|c|c|c|c|c}
		\toprule[1.5pt]
		Stage & Output & Operator & \#exp & \#out & SE & Stride \\ 
		\midrule
		\multirow{2}{*}{stem} & $112\times 112$   & Conv3$\times$3 &  -  & 16     &  -  & 2 \\
		& $112\times 112$   & G-bneck &  16  & 16     &  -  & 1 \\
		\midrule
		\multirow{2}{*}{1} & $56\times 56$   & G-bneck &  48  & 24     &  -  & 2 \\
		& $56\times 56$   & G-bneck &  72  & 24     &  -  & 1 \\
		\midrule
		\multirow{2}{*}{2} & $28\times 28$   & G-bneck &  72  & 40      &  1  & 2 \\
		& $28\times 28$   & G-bneck &  120  & 40     &  1  & 1 \\
		\midrule
		\multirow{6}{*}{3} & $14\times 14$   & G-bneck &  240  & 80     &  -  & 2 \\
		& $14\times 14$   & G-bneck &  200  & 80     &  -  & 1 \\
		& $14\times 14$   & G-bneck &  184  & 80    &  -  & 1 \\
		& $14\times 14$   & G-bneck &  184  & 80    &  -  & 1 \\
		& $14\times 14$   & G-bneck &  480  & 112     &  1  & 1 \\
		& $14\times 14$   & G-bneck &  672  & 112     &  1  & 1 \\
		\midrule
		\multirow{5}{*}{4} & $7\times 7$   & G-bneck &  672  & 160     &  1  & 2 \\
		& $7\times 7$   & G-bneck &  960  & 160      &  -  & 1 \\
		& $7\times 7$   & G-bneck &  960  & 160      &  1  & 1 \\
		& $7\times 7$   & G-bneck &  960  & 160      &  -  & 1 \\
		& $7\times 7$   & G-bneck &  960  & 160      &  1  & 1 \\
		\midrule
		\multirow{4}{*}{head} & $7\times 7$   & Conv1$\times$1 &  - & 960     &  -  & 1 \\
		& $1\times 1$   & AvgPool &  -  & 960   & -  & - \\
		& $1\times 1$   & Conv1$\times$1 &  -  & 1280     &  -  & 1 \\
		& $1\times 1$   & FC &  -  & 1000     &  -  & - \\
		\bottomrule[1pt]
	\end{tabular}
	\vspace{-1.0em}
\end{table}

\paragraph{C-GhostNet.}

Building on the C-Ghost bottleneck, we propose C-GhostNet as presented in Table~\ref{tab:C-GhostNet}. We basically follow the architecture of MobileNetV3~\citep{mobilenetv3} for its superiority and replace the bottleneck block in MobileNetV3 with our C-Ghost bottleneck. C-GhostNet mainly consists of a stack of C-Ghost bottlenecks with the C-Ghost modules as the building block. The first layer is a standard convolutional layer with 16 filters, then a series of C-Ghost bottlenecks with gradually increased channels are followed. These C-Ghost bottlenecks are grouped into different stages according to the sizes of their input feature maps. All the C-Ghost bottlenecks are applied with stride=1 except that the last one in each stage is with stride=2. At last a global average pooling and a convolutional layer are utilized to transform the feature maps to a 1280-dimensional feature vector for final classification. The squeeze and excite (SE) module~\citep{senet} is also applied to the residual layer in some ghost bottlenecks as in Table~\ref{tab:C-GhostNet}. In contrast to MobileNetV3, we do not use hard-swish nonlinearity function due to its large latency. The presented architecture provides a basic design for reference, although further hyper-parameters tuning or automatic architecture searching based ghost module will further boost the performance.

\paragraph{Width Multiplier.}
Although the given architecture in Table~\ref{tab:C-GhostNet} can already provide low latency and guaranteed accuracy, in some scenarios we may require smaller and faster models or higher accuracy on specific tasks. To customize the network for the desired needs, we can simply multiply a factor $\alpha$ on the number of channels uniformly at each layer. This factor $\alpha$ is called \emph{width multiplier} as it can change the width of the entire network. We denote C-GhostNet with width multiplier $\alpha$ as C-GhostNet-$\alpha\times$. Width multiplier can control the model size and the computational cost quadratically by roughly $\alpha^2$. Usually smaller $\alpha$ leads to lower latency and lower performance, and vice versa.

\section{GPU-Efficient GhostNet}
In this section, we describe the details of the proposed G-Ghost stage, a universal stage structure for building GPU-efficient network architectures.

\textbf{
	\subsection{G-Ghost Stage}
}
Although our C-GhostNets can save FLOPs while maintaining high performance, the used cheap operation for generating more features are still not very cheap and efficient on GPU. Specifically, the depth-wise convolution is usually of lower arithmetic intensity, \ie, ratio of computation to memory operations, which cannot be sufficiently utilized the parallel computing capability~\citep{roofline,squeezenext}. How to derive CNNs with better trade-off between accuracy and GPU latency is still a neglected problem. 

Besides the FLOPs and the number of parameters, Radosavovic et al. introduce \emph{activations} to measure the complexity of networks, \ie, the size of the output tensors of all convolutional layers~\citep{regnet}. The latency on GPUs is more relative to activations than FLOPs, that is to say, if we can remove part of feature maps and decrease activations, we can reduce the latency on GPUs with high probability. On the other hand, the main body of a CNN is usually composed of several stages with progressively reduced resolution, and each stage consists of a stack of blocks. Rather than block-wise enhancement which is investigated in previous works~\citep{mobilev2,squeezenet}, we target on reducing \emph{stage-wise} redundancy which can largely decrease the intermediate features and consequently cut down the related computational cost and memory usage.

The commonly-used backbone networks~\citep{alexnet,resnet,googlenet,efficientnet} usually consists of three parts: stem, body and head~\citep{regnet}. The bulk of computation and parameters are consumed by network body, while the stem and head are relatively lightweight. Thus, here we target to simplify the network body for efficient inference.
The network body is composed of several stages which process feature maps with progressively reduced resolution. For each stage in the CNN, we have $n$ layers (\eg, for AlexNet) or blocks (\eg, for ResNet), denoted as $\{L_1,L_2,\cdots,L_n\}$. Given the input feature maps $X$, the outputs of the first block and the last block are
\begin{align}
	Y_1 &= L_1(X),\\
	Y_n &= L_n(L_{n-1}(\cdots L_2(Y_1))),\label{eq:cnn}
\end{align}
respectively. To obtain the output feature $Y_n$, a sequence of blocks are executed to process and aggregate the input data, that is, large amount of computational cost is required.

\begin{figure}[htp]
	\centering
	\includegraphics[width=0.8\linewidth]{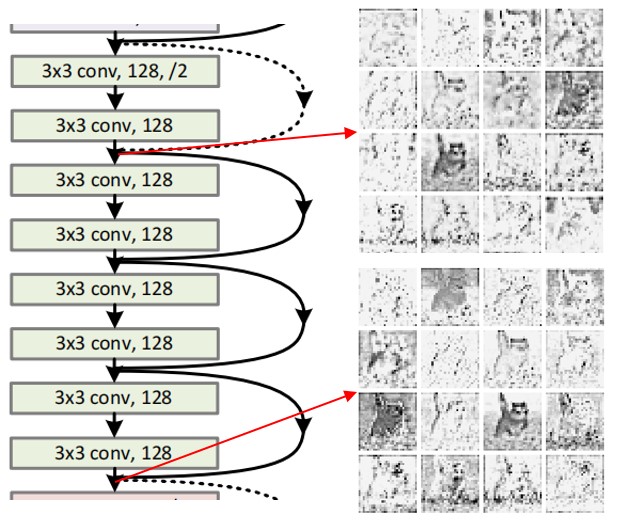}
	\vspace{-0.5em}
	\caption{Feature maps from the first block and the last block in the second stage of ResNet34.}
	\vspace{-0em}
	\label{Fig:feature-maps}
\end{figure}

\begin{figure*}[tp]
	\small
	\begin{center}
		\renewcommand{\arraystretch}{1.1}
		\setlength{\tabcolsep}{18pt}
		\begin{tabular}{ccc}			
			\includegraphics[height=0.22\textheight]{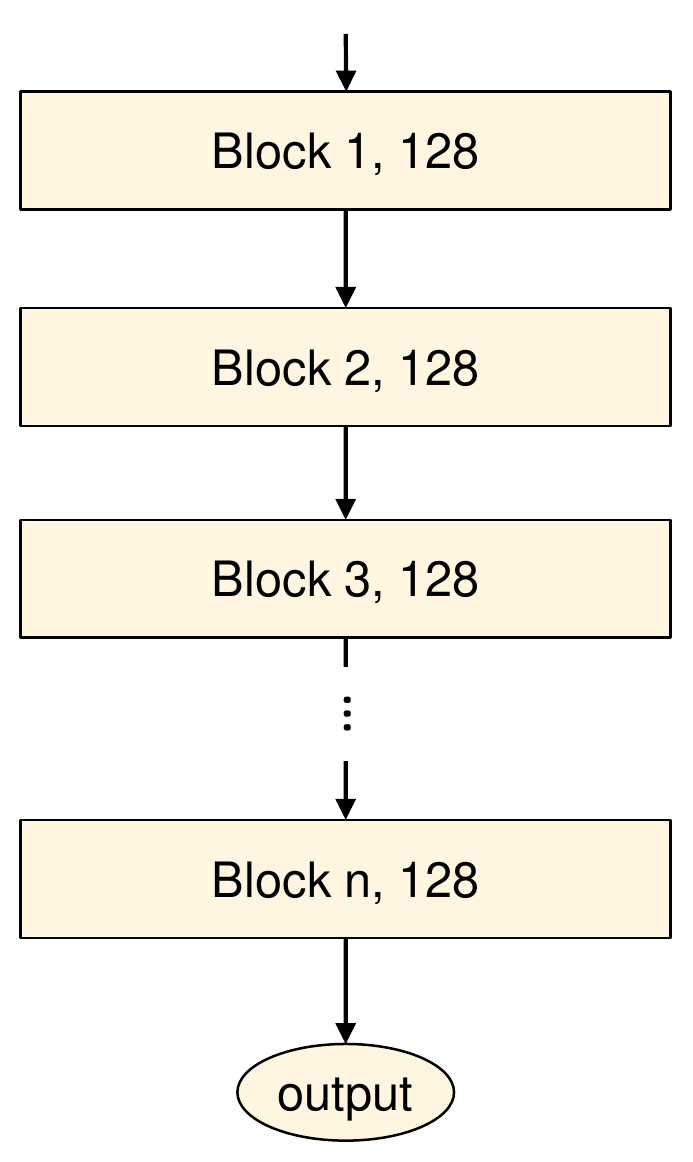}&
			\includegraphics[height=0.22\textheight]{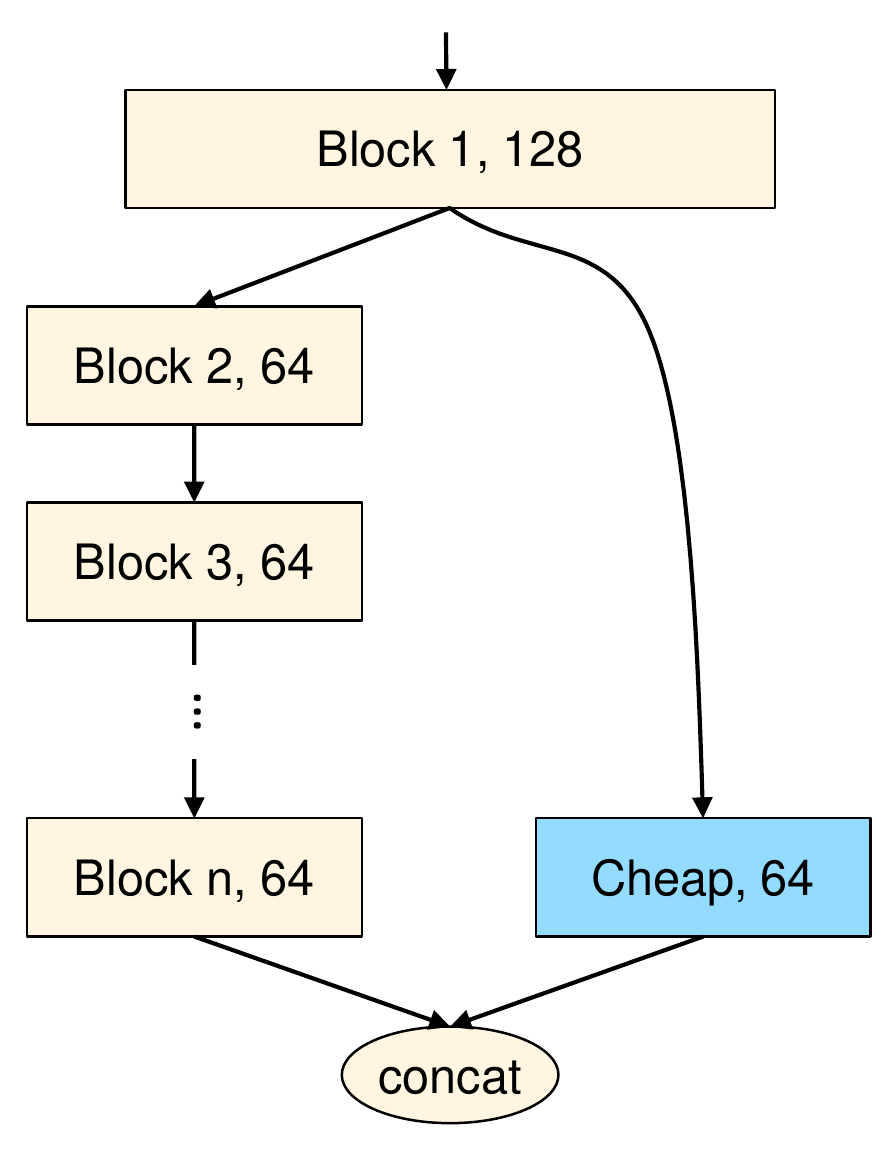}&
			\includegraphics[height=0.22\textheight]{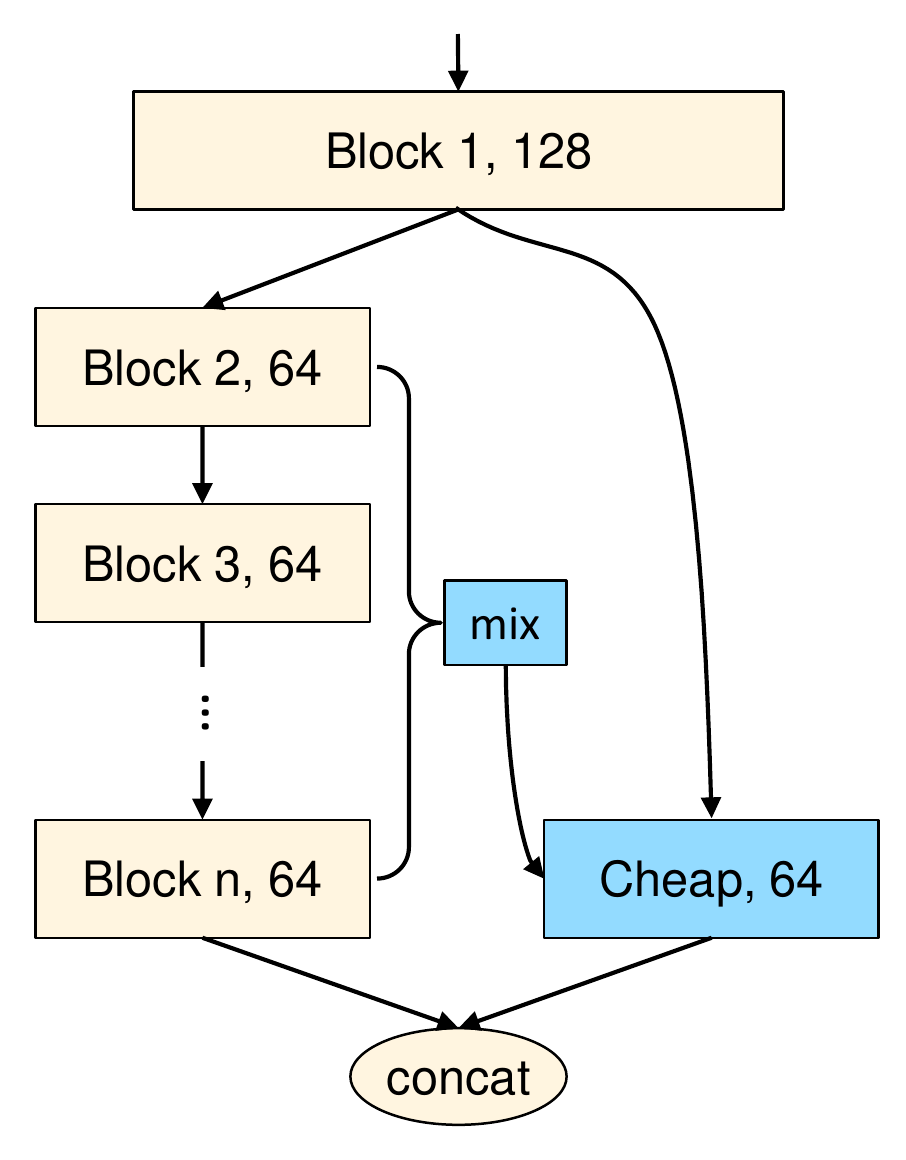}\\
			(a) Vanilla CNN stage. & (b) G-Ghost stage. & (c) G-Ghost stage w/ mix. \\
		\end{tabular}
		\vspace{-0.5em}
		\caption{Illustration of vanilla CNN stage and the proposed G-Ghost stage, where the block can be any type of block, \eg, residual block~\citep{resnet}, inverted residual block~\citep{mobilev2} and ghost block.}
		\label{Fig:G-Ghost}
	\end{center}
	\vspace{-1.em}
\end{figure*}

We argue that not all the features $Y_n$ from the last block need to be produced using such expensive computation of multiple blocks. In Figure~\ref{Fig:feature-maps}, we show the feature maps from the first block and the last block in the third stage of ResNet34. Although the features of the last block are processed by 5 more blocks, some of them are very similar to those of the first block, which means these features can be obtained by simple transformation from the low-level features.

In general, we divide the deep features into \emph{complicated} and \emph{ghost} ones, where the complicated features still require a sequence of blocks to generate, while the ghost ones can be easier to obtain from the shallow features. For a stage of $n$ blocks with output features $X\in\mathbb{R}^{c\times h\times w}$ where $c$, $h$, $w$ are the number of channels, the height and the width, respectively, we denote its complicated features as $Y_n^c\in\mathbb{R}^{(1-\lambda)c\times h\times w}$ and ghost features as $Y_n^g\in\mathbb{R}^{\lambda c\times h\times w}$, where $0\leq\lambda\leq1$ is the ratio of ghost features. The complicated features are produced with $n$ blocks:
\begin{equation}\label{eq:complicated}
	Y_n^c = L'_n(L'_{n-1}(\cdots L'_2(Y_1))),
\end{equation}
where the blocks $L'_2,\cdots,L'_n$ are thin blocks as Eq.~\ref{eq:cnn} with $(1-\lambda)\times$ width. As analyzed in the above, the simple features $Y_n^g$ are similar to features $Y_1$ of the first block, which can be viewed as ghosts of $Y_1$. We transform the features $Y_1$ to generate $Y_n^g$ with cheap operation $C$:
\begin{equation}\label{eq:cheap}
	Y_{n}^g = C(Y_1),
\end{equation}
where cheap operation $C$ can simply be 1$\times$1 convolution, 3$\times$3 convolution, \etc. By concatenating $Y_n^c$ and $Y_{n}^g$, we can obtain the final output of the stage:
\begin{equation}\label{eq:concat}
	Y_n = [Y_n^c, Y_{n}^g].
\end{equation}
The proposed structure as in Eq.~\ref{eq:complicated}-\ref{eq:concat} is called G-Ghost stage, and Figure~\ref{Fig:G-Ghost}(b) gives an illustration of G-Ghost stage with $\lambda=0.5$.

\textbf{
\subsection{Intrinsic Feature Aggregation}
}
The proposed G-Ghost stage generates parts of features with cheap operations by exploiting the redundancy between the first block and the last block. In this way, the computational cost can be reduced largely compared to vanilla CNN stage (Figure~\ref{Fig:G-Ghost}(a)). Although the simple features can be approximately generated by cheap operations, $Y_{n}^2$ in Eq.~\ref{eq:cheap} may lack deep information which need multiple layers to extract. To complement the lacked information, we propose to utilize the intermediate features in the complicated path to enhance the representation capacity of the cheap operations. 

\begin{figure}[htp]
	\centering
	\includegraphics[width=0.75\linewidth]{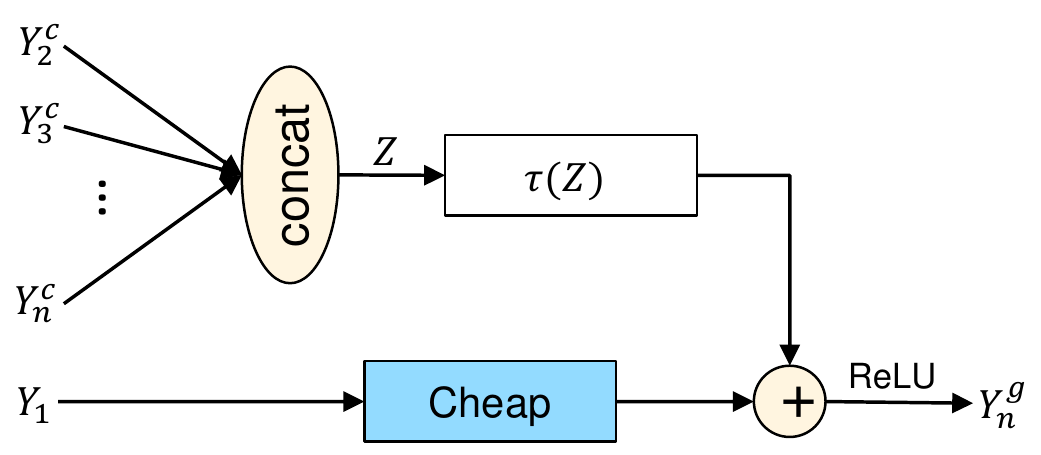}
	\caption{Mix operation for aggregating the intermediate information into the ghost features.}
	\label{Fig:mix}
	\vspace{-1.em}
\end{figure}

From the complicated path in G-Ghost stage, we can collect the intermediate features $Z\in\mathbb{R}^{c'\times h\times w}=[Y_2^c,Y_3^c,\cdots,Y_n^c]$ where $c'$ is the total number of channels. These intermediate features extracted from multiple layers can provide rich information supplementation to the cheap operation. As shown in Figure~\ref{Fig:mix}, we first transform $Z$ into the same domain as $Y_{n}^g$, and then mix the transformed information into $Y_{n}^2$:
\begin{equation}\label{eq:mix}
	Y_{n}^g \leftarrow Y_{n}^g + \tau(Z),
\end{equation}
where $\tau$ is the transformation function. In order not to bring in much extra computation, we keep the transformation $\tau$ simple. It may have various candidates, and here we supply a simple design. We first apply a global average pooling to obtain the aggregated feature $z\in\mathbb{R}^{c'}=\emph{Pooling}(Z)$. A fully connected layers are applied to transform $z$ into the same domain as $Y_{n}^2$:
\begin{equation}\label{eq:fc}
	\tau(Z) = W\textit{Pooling}(Z)+b,
\end{equation}
where $W\in\mathbb{R}^{c'\times \lambda c}$ and $b\in\mathbb{R}^{\lambda c}$ are weight and bias respectively. As shown in Figure~\ref{Fig:G-Ghost}(c), the mix operation is applied in G-Ghost stage for intrinsic feature aggregation.

\paragraph{Discussion with Existing Methods.} 
The proposed G-Ghost stage and the existing efficient block design (\eg, CondenseNet V2~\citep{yang2021condensenet}) both utilize the redundancy in feature maps to reduce the complexity of CNNs. They have many differences, mainly including 1) CondenseNet V2 focuses on the channel redundancy in the same layer, while G-Ghost reduces the redundancy across different layers or blocks; 2) CondenseNet V2 introduces a sparse feature reactivation (SFR) module for feature reusing in a block, while G-Ghost proposes a new structure to improve the CNN stage structure; 3) CondenseNet V2 adopts sparse connection in each SFR module which is unfriendly for memory-bound devices, while G-Ghost uses ordinary convolution and reduces activations effectively, resulting in faster inference on GPUs. 4) The mix operation in G-Ghost utilizes a global pooling (with a FC layer) to aggregate the information of intermediate layers. CondenseNet V2 utilizes a learned sparse connection to select important features from preceding layers.

Compared to the existing efficient network architecture (\eg, DenseNet \citep{densenet}), the proposed G-Ghost stage mainly differs at 1) the existing DenseNet is a specific network architecture, while G-Ghost stage is a general structure which can be applied to simplify most of CNNs; 2) The feature dimension in DenseNet increase as the layer goes deeper, while G-Ghost reduces the feature dimension in the intermediate layers and keeps the output as same as the original network; 4) In DenseNet, feature maps from previous blocks are concatenated to the output of each layer, while G-Ghost utilizes a cheap operation to transform the feature maps and concatenate them to the final output of the stage (not each layer); 4) We introduce a mix operation in G-Ghost stage to aggregate the information of intermediate layers.

\textbf{
\subsection{Complexity Analysis}
}
By upgrading the original CNN structure with the proposed G-Ghost stage, the computation and memory cost can be largely reduced. The reduction ratio is related to the original stage structure, the ghost ratio $\lambda$, the cheap operation and the mix operation. Given a stage with $n$ blocks, the FLOPs of the $i$th blocks are $f_i$, and the parameters are $p_i$ ($i=1,\cdots,n$), respectively. G-Ghost stage reduces the FLOPs of the original $n$ blocks to $f_1$, $(1-\lambda)f_2$, and $\{(1-\lambda)^2f_i\}_{i=3}^{n}$, respectively. The parameters are reduced to $p_1$, $(1-\lambda)p_2$, and $\{(1-\lambda)^2p_i\}_{i=3}^{n}$, respectively. The introduced cheap operation has $f^c$ FLOPs and $p^c$ parameters. The reduction ratio of FLOPs and parameters are
\begin{align}
	r_f &\approx \frac{\sum_{i=1}^{n}f_n}{f_1+(1-\lambda)f_2+\sum_{i=3}^{n}(1-\lambda)^2f_n+f^c},\label{eq:ratio1}\\
	r_p &\approx \frac{\sum_{i=1}^{n}p_n}{p_1+(1-\lambda)p_2+\sum_{i=3}^{n}(1-\lambda)^2p_n+p^c},\label{eq:ratio2}
\end{align}
respectively. It can be seen from Eq.~\ref{eq:ratio1} and Eq.~\ref{eq:ratio2} that the FLOPs or parameters reduction is appreciable when $n\geq3$, so we recommend to use G-Ghost stage in the stages with more than 2 blocks. In general, the larger ghost ratio $\lambda$ leads to larger reduction ratio. If the ghost ratio is too large, the representation capacity of the network will be much weaker, so we need to take a trade-off between the computational cost and the performance.


\begin{table}[htb]
	\vspace{-1em}
	\centering
	\renewcommand\arraystretch{1.0}
	\caption{Overall architecture of G-GhostNet. Block denotes residual bottleneck. output means the output spatial size. \#out means the number of output channels.}
	\label{tab:G-GhostNet}
	\small
	\setlength{\tabcolsep}{7pt}{
		\begin{tabular}{c|c|cl|c|c}
			\toprule[1.5pt]
			Stage & Output & \multicolumn{2}{c|}{Operator} & \#out & Stride \\
			\midrule
			stem & $112\times 112$   & \multicolumn{2}{c|}{Conv3$\times$3}  & 16     &  2 \\
			\midrule
			\multirow{3}{*}{1} & $56\times 56$   & \multicolumn{2}{c|}{Block}  & 24     &  2 \\
			& $56\times 56$   & Block$\times$1 &  Cheap  & 24      &  1 \\
			& $56\times 56$   & \multicolumn{2}{c|}{Concat}  & 24      &  1 \\
			\midrule
			\multirow{3}{*}{2} & $28\times 28$   & \multicolumn{2}{c|}{Block}  & 40     &  2 \\
			& $28\times 28$   & Block$\times$1 &  Cheap  & 40     &  1 \\
			& $28\times 28$   & \multicolumn{2}{c|}{Concat}  & 40     &  1 \\
			\midrule
			\multirow{3}{*}{3} & $14\times 14$   & \multicolumn{2}{c|}{Block}  & 80     &  2 \\
			& $14\times 14$   & Block$\times$5 &  Cheap  & 80    &  1 \\
			& $14\times 14$   & \multicolumn{2}{c|}{Concat}  & 80    &  1 \\
			\midrule
			\multirow{3}{*}{4} & $7\times 7$   & \multicolumn{2}{c|}{Block}  & 160      &  2 \\
			& $7\times 7$   & Block$\times$5 &  Cheap  & 160      &  1 \\
			& $7\times 7$   & \multicolumn{2}{c|}{Concat}  & 160      &  1 \\
			\midrule
			\multirow{3}{*}{head} & $7\times 7$   & \multicolumn{2}{c|}{Conv1$\times$1}  & 960     &  1 \\
			& $1\times 1$   &  \multicolumn{2}{c|}{Pool \& Conv1$\times$1}  & 1280      &  1 \\
			& $1\times 1$   & \multicolumn{2}{c|}{FC}  & 1000     &  - \\
			\bottomrule[1pt]
		\end{tabular}
	}
	\vspace{-1em}
\end{table}

\textbf{
\subsection{Building Lightweight G-GhostNet}
}
We can utilize the proposed G-Ghost stage to remould the existing CNN architectures (\eg, ResNet) by replacing the vanilla stage with G-Ghost stage. By exploiting stage-wise representation redundancy, G-Ghost stage can achieve a better trade-off between accuracy and GPU latency.

We also build a new lightweight network using G-Ghost stage, named G-GhostNet. The aforementioned C-GhostNet in the last section consisting of massive depth-wise convolutions is designed under FLOPs constraint and is unfriendly to GPUs. To develop a lightweight model for efficient inference on GPU devices, we build the G-GhostNet using our G-Ghost stage with mix operation. We follow the basic structure of C-GhostNet and remould all the stages with G-Ghost stage as shown in Table~\ref{tab:G-GhostNet}. In all the G-Ghost stages, the 1$\times$1 convolution is used as the cheap operation and the ghost ratio $\lambda$ is set to 0.4. The expansion ratio in the residual bottleneck is set as 3, and SE module~\citep{senet} is applied on each block. The simple ReLU is adopted as the activation function for its efficiency. We denote G-GhostNet-$\alpha\times$ as the G-GhostNet model with width multiplier of $\alpha$.

\vspace{1em}
\section{Experiments}\label{Experiments}
In this section, we conduct extensive experiments to verify the efficiency of the proposed C-Ghost module on CPU and G-Ghost stage on GPU.

\textbf{\subsection{Datasets and Experimental Settings}}
To verify the effectiveness of the proposed C-Ghost module and C-GhostNet architecture, we conduct experiments on several benchmark visual datasets, including CIFAR-10~\citep{cifar}, ImageNet ILSVRC 2012 dataset~\citep{imagenet}, and MS COCO object detection benchmark~\citep{coco}.

\paragraph{CIFAR10.} CIFAR10 dataset consists of 60000 32$\times$32 colour images in 10 classes, with 6000 images per class. There are 50000 training images and 10000 test images~\citep{cifar}. Data augmentation including random crop and random flip is applied.

\paragraph{ImageNet.} ImageNet ILSVRC2012 is a large-scale image classification dataset, containing the 1000 categories and 1.2 million images. A subset of 50,000 of the images with labels are used as validation data~\citep{imagenet}. The commonly-used random crop, random clip and color jitter strategies are used for data augmentation~\citep{inceptionv3}.

\paragraph{MS COCO.} We evaluate the generalization of our method on COCO 2014~\citep{coco} detection task. We train models on COCO \emph{trainval35k} split (union of 80K training images and 35K subset of images from validation set) and evaluate on the \emph{minival} split with 5K images~\citep{fpn}. The input image is resized to have an 800-pixel shorter side.

\paragraph{Implementation Details.} All the models are implemented using PyTorch~\citep{pytorch} and trained on NVIDIA V100 GPUs. In CIFAR10 experiments, the training settings basically follow the common practice~\citep{wresnet,fpgm}. For ImageNet, we keep the training scheme the same as the baseline networks for fair comparison. ResNet models are trained for 100 epochs with hype-parameters in Pytorch official examples~\citep{pytorch}. The settings for RegNet models are the same as those in the original paper~\citep{regnet}. The C-GhostNet and G-GhostNet models in mobile setting are trained using the training setup as in~\citep{mobilenetv3}. We do not use advanced data augmentation, \eg, RandAugment~\citep{randaugment}, unless specifically noted otherwise. As for object detection experiments, we use MMDetection~\citep{mmdetection} framework to implement our models and train them under the common 1$\times$ scheduler setup.

\textbf{
	\subsection{CPU-Efficient C-GhostNet}
}
In this section, we first replace the original convolutional layers by the proposed C-Ghost module to verify its effectiveness. Then, the C-GhostNet architecture built using the new module will be further tested on the image classification and object detection benchmarks.

\subsubsection{Toy Experiments} We have presented a diagram in Figure~\ref{Fig:maps} to point out that there are some similar feature map pairs, which can be efficiently generated using some efficient linear operations. Here we first conduct a toy experiment to observe the reconstruction error between raw feature maps and the generated ghost feature maps. Taking three pairs in Figure~\ref{Fig:maps} (\ie, red, greed, and blue) as examples, features are extracted using the first residual block of ResNet50~\citep{resnet}. Taking the feature on the left as input and the other one as output, we utilize a small depth-wise convolution filter to learn the mapping, \ie, the linear operation $\Phi$ between them. The size of the convolution filter $d$ is ranged from $1$ to $7$, MSE (mean squared error) values of each pair with different $d$ are shown in Table~\ref{tab:toy}.

\begin{table}[htb]
	\centering
	\small
	\renewcommand\arraystretch{1.0}
	\caption{MSE error \emph{v.s.} different kernel sizes.}
	\vspace{0.1em}
	\label{tab:toy}	
	\setlength{\tabcolsep}{8pt}{
	\begin{tabular}{l|c|c|c|c}
		\toprule[1.5pt]
		MSE ($10^{-3}$) & $d$=1 & $d$=3 & $d$=5 & $d$=7 \\
		\midrule
		red pair &  $4.0$    &  $3.3$    & $3.3$ &  $3.2$    \\
		green pair &  $25.0$     &  $24.3$    & $24.1$ &  $23.9$    \\
		blue pair &  $12.1$     &  $11.2$    & $11.1$ &  $11.0$    \\
		\bottomrule[1pt]
	\end{tabular}
	}
	\vspace{-1em}
\end{table}

It can be found in Table~\ref{tab:toy} that all the MSE values are extremely small, which demonstrates that there are strong correlations between feature maps in deep neural networks and these redundant feature maps could be generated from several intrinsic feature maps. The similarity phenomenon holds for most of images, as shown in this link: \url{https://bit.ly/38jJtR6} (ResNet50-Group1 feature maps of randomly sampled 1000 ImageNet images). The toy experiment results hold for these images too. Besides convolutions used in the above experiments, we can also explore some other low-cost linear operations to construct the C-Ghost module such as affine transformation and wavelet transformation. However, convolution is an efficient operation already well support by current hardware and it can cover a number of widely used linear operations such as smoothing, blurring, and motion. Moreover, although we can also learn the size of each filter \wrt the linear operation $\Phi$, the irregular module will reduce the efficiency of computing units (\eg, CPU). Thus, we suggest to let $d$ in a C-Ghost module be a fixed value and utilize depth-wise convolution to implement Eq.~\ref{eq:ghost} for building highly efficient deep neural networks in the following experiments.

\begin{table}[htb]
	\vspace{-0em}
	\centering
	\small
	\renewcommand\arraystretch{1.0}
	\caption{The performance of the proposed C-Ghost module with different $d$ on CIFAR-10.}
	\vspace{0.1em}
	\label{tab:p}
	\begin{tabular}{c|c|c|c}
		\toprule[1.5pt]
		$d$         &  Params (M) & FLOPs (M) & Acc (\%) \\
		\midrule
		VGG16         & 15.0    & 313    & 93.6     \\ 
		1         & 7.6   & 157  & 93.5     \\ 
		3       &    7.7  & 158 &    93.7 \\
		5       &   7.7  & 160   &  93.4  \\ 
		7        &   7.7  & 163    & 93.1  \\
		\bottomrule[1pt]
	\end{tabular}
	\vspace{-1em}
\end{table}

\begin{table}[htb]
	\vspace{-0em}
	\centering
	\small
	\renewcommand\arraystretch{1.0}
	\caption{The performance of the proposed C-Ghost module with different $s$ on CIFAR-10.}
	\vspace{0.1em}
	\label{tab:s}
	\begin{tabular}{c|c|c|c}
		\toprule[1.5pt]
		$s$      &  Params (M)    &  FLOPs (M) & Acc (\%) \\
		\midrule
		VGG16      & 15.0   & 313        & 93.6     \\ 
		2     & 7.7     & 158        & 93.7     \\ 
		3     & 5.2 & 107 &  93.4 \\
		4      & 4.0  &   80      &  93.0   \\
		5     & 3.3   &   65      &  92.9  \\
		\bottomrule[1pt]
	\end{tabular}
	\vspace{-1em}
\end{table}

\subsubsection{CIFAR-10 Experiments} We evaluate the proposed C-Ghost module on two popular network architectures, \ie, VGG16~\citep{vgg} and ResNet56~\citep{resnet}, on CIFAR-10 dataset. Since VGG16 is originally designed for ImageNet, we use its variant~\citep{cifar-vgg} which is widely used in literatures for conducting the following experiments. All the convolutional layers in these two models are replaced by the proposed C-Ghost module, and the new models are denoted as C-Ghost-VGG16 and Ghost-ResNet56, respectively. Our training strategy closely follows the settings in~\citep{resnet}, including momentum, learning rate, \etc. We first analyze the effects of the two hyper-parameters $s$ and $d$ in C-Ghost module, and then compare the Ghost-models with the state-of-the-art methods.

%

\paragraph{Analysis on Hyper-parameters.} As described in Eq.~\ref{eq:ghost}, the proposed Ghost Module for efficient deep neural networks has two hyper-parameters, \ie, $s$ for generating $m = n/s$ intrinsic feature maps, and kernel size $d\times d$ of linear operations (\ie, the size of depth-wise convolution filters) for calculating ghost feature maps. The impact of these two parameters are tested on the VGG16 architecture.

First, we fix $s=2$ and tune $d$ in $\{1,3,5,7\}$, and list the results on CIFAR-10 validation set in Table~\ref{tab:p}. We can see that the proposed C-Ghost module with $d=3$ performs better than smaller or larger ones. This is because that kernels of size $1\times 1$ cannot introduce spatial information on feature maps, while larger kernels such as $d=5$ or $d=7$ lead to overfitting and more computations. Therefore, we adopt $d=3$ in the following experiments for effectiveness and efficiency.

After investigating the kernel sizes used in the proposed C-Ghost module, we keep $d=3$ and tune the other hyper-parameter $s$ in the range of $\{2,3,4,5\}$. In fact, $s$ is directly related to the computational costs of the resulting network, that is, larger $s$ leads to larger compression and  speed-up ratio as analyzed in Eq.~\ref{eq:rc} and Eq.~\ref{eq:rs}. From the results in Table~\ref{tab:s}, when we increase $s$, the FLOPs are reduced significantly and the accuracy decreases gradually, which is as expected. Especially when $s=2$ which means compress VGG16 by $2\times$, our method performs even slightly better than the original model, indicating the superiority of the proposed C-Ghost module.

\begin{table}[htb]
	\centering
	\small
	\renewcommand\arraystretch{1.0}
	\caption{Comparison of state-of-the-art methods for compressing VGG16 and ResNet56 on CIFAR-10. - represents no reported results available.}
	\vspace{0.1em}
	\label{tab:cifar10}
	\setlength{\tabcolsep}{5pt}{
	\begin{tabular}{l|c|c|c}
		\toprule[1.5pt]
		Model         & Params (M) & FLOPs (M) & Acc (\%) \\
		\midrule
		VGG16   & 15        & 313        & 93.6     \\ 
		PFEC~\citep{l1-pruning} & 5.4 & 206 & 93.4 \\
		CP~\citep{cp} & - & 136 & 92.5 \\
		C-Ghost ($s$=2) & 7.7         &   158      & \textbf{93.7}     \\ 
		\midrule
		ResNet56   & 0.85         & 125        & 94.1     \\
		CP~\citep{cp} & - & 63 & 92.0 \\ 
		PFEC~\citep{l1-pruning} & 0.73 & 91 & 92.5 \\
		AMC~\citep{amc} & - & 63 & 91.9 \\
		GAL~\citep{gal} & 0.75 & 78 & 93.0 \\
		HRank~\citep{hrank} & 0.49 & 63 &  93.2 \\
		C-Ghost ($s$=2) & 0.43     & 63   & \textbf{93.8}     \\
		\bottomrule[1pt]
	\end{tabular}
	}
	\vspace{-1em}
\end{table}

\begin{table*}[htb]
	\centering
	\small
	\renewcommand\arraystretch{1.0}
	\caption{Comparison of state-of-the-art methods for compressing ResNet50 on ImageNet dataset. The CPU latency is measured on an Intel CPU where - denotes no public code available.}
	\label{tab:resnet50}
	\setlength{\tabcolsep}{8pt}{
	\begin{tabular}{l|c|c|c|c|c}
		\toprule[1.5pt]
		Model         & Params (M) & FLOPs (B) & CPU Latency (ms) & Top-1 (\%) & Top-5 (\%) \\
		\midrule
		ResNet50~\citep{resnet}   & 25.6     & 4.1   & 158      & 76.1 &  92.8    \\
		\midrule
		NISP-B~\citep{nisp} & 14.4 & 2.3 & -  & - & 90.8 \\ 
		Versatile Filters~\citep{versatile} & 11.0 &  3.0 & 172  & 74.5 & 91.8 \\
		SSS~\citep{huang2018data} & - & 2.8 & -  & 74.2 & 91.9 \\
		GAL-0.5~\citep{gal} & 21.2 & 2.3 & -  & 72.0 & 90.9 \\
		PFP-B~\citep{liebenwein2019provable} & - & 2.9 & -  & 75.2 & 92.4 \\
		DSA~\citep{ning2020dsa} & - & 2.1 & 174  & 74.7 & 92.1 \\
		C-Ghost ($s$=2) & 13.0      & 2.2   & 129    &  \textbf{75.7} & \textbf{92.6}     \\
		\midrule
		Shift~\citep{shift} & 6.0 & - & -  & 70.6 & 90.1  \\
		Taylor-FO-BN~\citep{molchanov2019importance} & 7.9 & 1.3 & -  & 71.7 & - \\
		Slimmable 0.5$\times$~\citep{slimmable} & 6.9 & 1.1 & 160  & 72.1 & -  \\
		HRank~\citep{hrank} & 13.8 & 1.6 & 120  & 72.0 & 91.0 \\
		MetaPruning~\citep{metapruning} & - & 1.0 & 78  & 73.4 & - \\
		LFPC~\citep{he2020learning} & - & 1.6 & 119  & 74.5 & 92.0 \\
		C-Ghost ($s$=4) & 6.5   & 1.2  & 112  & \textbf{74.9} & \textbf{92.3}  \\
		\bottomrule[1pt]
	\end{tabular}
	}
	\vspace{-1em}
\end{table*}

\paragraph{Comparison with State-of-the-arts.} We compare Ghost-Net with several representative state-of-the-art models on both VGG16 and ResNet56 architectures. The compared methods include different types of model compression approaches, PFEC~\citep{l1-pruning,rethinking-pruning}, channel pruning (CP)~\citep{cp}, AMC~\citep{amc} and HRank~\citep{hrank}. For VGG16, our model can obtain an accuracy slightly higher than the original one with a 2$\times$ acceleration, which indicates that there is considerable redundancy in the VGG model. Our C-Ghost-VGG16 ($s=2$) outperforms the competitors with the highest performance ($93.7\%$) but with significantly fewer FLOPs. For ResNet56 which is already much smaller than VGG16, our model can achieve comparable accuracy with baseline with 2$\times$ speed-up. We can also see that other state-of-the-art models with similar or larger computational cost obtain lower accuracy than ours.

\begin{figure}[htb]
	\vspace{-0em}
	\centering
	\includegraphics[width=1.0\linewidth]{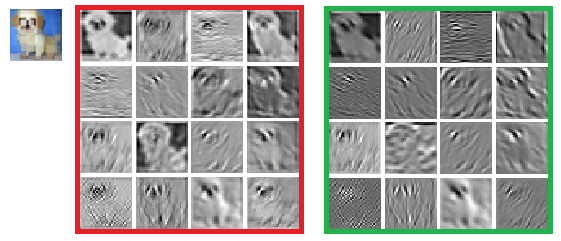}
	\caption{The feature maps in the 2nd layer of C-Ghost-VGG16. The left-top image is the input, the feature maps in the left red box are from the primary convolution, and the feature maps in the right green box are after the depth-wise transformation.}
	\label{fig:vis}
	\vspace{-1em}
\end{figure}

\paragraph{Visualization of Feature Maps.} 
We also visualize the feature maps of our ghost module as shown in Figure~\ref{fig:vis}. Although the generated feature maps are from the primary feature maps, they exactly have significant difference which means the generated features are flexible enough to satisfy the need for the specific task.

\begin{figure}[htb]
	\centering
	\includegraphics[width=0.85\linewidth]{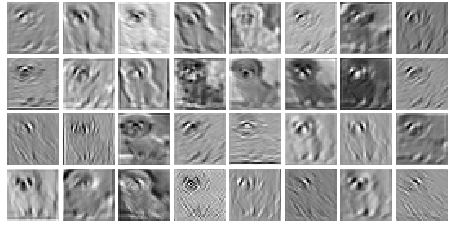}
	\caption{The feature maps in the 2nd layer of vanilla VGG16.}
	\label{fig:vgg-vis}
	\vspace{-1em}
\end{figure}


\subsubsection{Large Models on ImageNet} 
We next embed the C-Ghost module in the standard ResNet50 \citep{resnet} and conduct experiments on the large-scale ImageNet dataset. ResNet50 has about 25.6M parameters and 4.1B FLOPs with a top-1 accuracy of $76.1\%$. We use our C-Ghost module to replace all the convolutional layers in ResNet50 to obtain compact models and compare the results with several state-of-the-art methods, as detailed in Table~\ref{tab:resnet50}. The training settings such as the optimizer, learning rate, and batch size, are totally the same as those in Pytorch examples~\citep{pytorch} for fair comparisons.

From the results in Table~\ref{tab:resnet50}, we can see that our Ghost-ResNet50 ($s$=2) obtains about 2$\times$ acceleration and compression ratio, while maintaining the accuracy as that of the original ResNet50. Compared with the recent state-of-the-art methods including NISP~\citep{nisp}, Versatile filters~\citep{versatile}, Sparse structure selection (SSS)~\citep{huang2018data} and HRank~\citep{hrank}, our method can obtain significantly better performance under the 2$\times$ acceleration setting. When we further increase $s$ to 4, Ghost-based model has only a $1.2\%$ accuracy drop with an about 4$\times$ computation speed-up ratio. In contrast, compared methods~\citep{hrank,slimmable} with similar weights or FLOPs have much lower performance than ours.

\begin{table*}[htb]
	\centering
	\small
	\renewcommand\arraystretch{1.0}
	\caption{Comparison of state-of-the-art small networks over classification accuracy, the number of weights and FLOPs on ImageNet dataset. Several models' latency is not reported because shuffle operation is not supported in TFLite or no public code is available.}
	\vspace{0.1em}
	\label{tab:C-GhostNet-imagenet}
	\begin{tabular}{l|c|c|c|c|c}
		\toprule[1.5pt]
		Model         & Params (M) & FLOPs (M) & ARM Latency (ms) & Top-1 (\%) & Top-5 (\%) \\
		\midrule
		MobileNetV2 0.35$\times$~\citep{mobilev2}   &  1.7   &  59  &  14.5   &  60.3  & 82.9  \\
		ShuffleNetV2 0.5$\times$~\citep{shufflev2}   & 1.4   & 41 &  -    & 61.1  & 82.6  \\
		MobileNetV3 Large 0.35$\times$~\citep{mobilenetv3}   & 2.2  &  40  &  11.8   &  64.2   & -  \\
		CondenseNetV2-A~\citep{yang2021condensenet} & 2.0 & 46 &  - & 64.4 & 84.2 \\
		C-GhostNet 0.5$\times$ &  2.6   &  42 &  10.8  &  \textbf{66.2}   & \textbf{86.6}   \\ 
		\midrule
		MobileNetV1 0.5$\times$~\citep{mobilenet}   &  1.3   &  150   &  19.7   &  63.3  & 84.9   \\
		ShuffleNetV1 1.0$\times$ (g=3)~\citep{shufflenet}   & 1.9   & 138    &  -  & 67.8  & 87.7   \\
		MobileNetV2 0.75$\times$ (192$\times$192)~\citep{mobilev2}   & 2.6   &  153  &  28.6   & 68.7  & -\\
		ShuffleNetV2 1.0$\times$~\citep{shufflev2}   & 2.3   & 146  &  -    & 69.4  & 88.9  \\
		MobileNetV3 Large 0.75$\times$~\citep{mobilenetv3}   & 4.0  &  155   &  32.2   &  73.3   & -  \\
		C-GhostNet 1.0$\times$ & 5.2    & 141  & 31.1 &  \textbf{73.9}  & \textbf{91.4}   \\ 
		\midrule
		MobileNetV2 1.0$\times$~\citep{mobilev2}   & 3.5   & 300   &  50.5   & 71.8  & 91.0  \\
		ShuffleNetV2 1.5$\times$~\citep{shufflev2}   & 3.5   & 299   &  -   & 72.6  & 90.6   \\
		FE-Net 1.0$\times$~\citep{sparse-shift}   & 3.7   & 301  &  -    & 72.9  & -   \\
		FBNet-B~\citep{fbnet}    & 4.5   &  295  &  53.8  & 74.1  & -   \\
		MobileNeXt 1.0$\times$~\citep{mobilenext} & 3.4 & 300 &  54.9 & 74.0 & - \\
		ProxylessNAS~\citep{proxylessnas} & 4.1  &  320   &  52.5   &  74.6   &  92.2  \\
		MnasNet-A1~\citep{mnasnet}   & 3.9  &  312   &  76.2   &  75.2   &  92.5  \\
		MobileNetV3 Large 1.0$\times$~\citep{mobilenetv3}   & 5.4  &  219  &  46.4   &  75.2   & -  \\
		C-GhostNet 1.3$\times$ &  7.3    & 226  &  46.9  &  \textbf{75.7}  & \textbf{92.7}   \\ 
		\bottomrule[1pt]
	\end{tabular}
	\vspace{-1em}
\end{table*}

\subsubsection{CPU-Efficient C-GhostNet}
After demonstrating the superiority of the proposed C-Ghost module for efficiently generating feature maps, we then evaluate the well designed C-GhostNet architecture as shown in Table~\ref{tab:C-GhostNet} using C-Ghost bottlenecks on image classification and object detection tasks, respectively.

\paragraph{ImageNet Classification} To verify the superiority of the proposed C-GhostNet, we conduct experiments on ImageNet classification task. We follow most of the training settings used in~\citep{shufflenet}, except that the initial learning rate is set to 0.4 when batch size is 1,024 on 8 GPUs. All the results are reported with single crop top-1 performance on ImageNet validation set. For C-GhostNet, we set kernel size $k=1$ in the primary convolution and $s=2$ and $d=3$ in all the C-Ghost modules for simplicity.

Several modern small network architectures are selected as competitors, including MobileNet series~\citep{mobilenet,mobilev2,mobilenetv3}, ShuffleNet series~\citep{shufflenet,shufflev2}, ProxylessNAS~\citep{proxylessnas}, FBNet~\citep{fbnet}, MnasNet~\citep{mnasnet}, \etc. The results are summarized in Table~\ref{tab:C-GhostNet-imagenet}. The models are grouped into three levels of computational complexity typically for mobile applications, \ie, $\sim$50, $\sim$150, and 200-300 MFLOPs. From the results, we can see that generally larger FLOPs lead to higher accuracy in these small networks which shows the effectiveness of them. Our C-GhostNet outperforms other competitors consistently at various computational complexity levels, since C-GhostNet is more efficient in utilizing computation resources for generating feature maps.

\begin{figure}[htb]
	\vspace{-0.5em}
	\centering
	\small
	\includegraphics[width=0.9\linewidth]{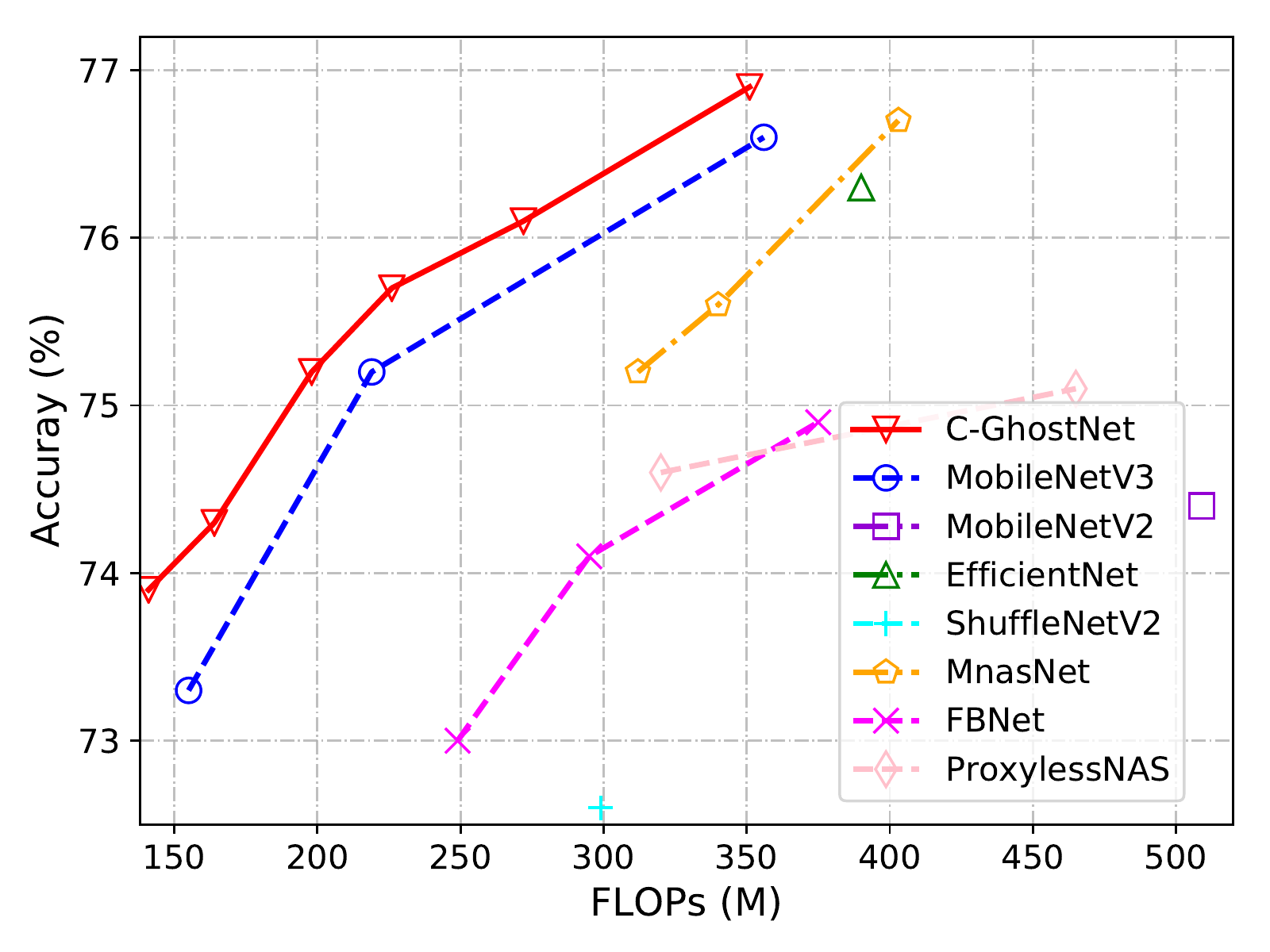}
	\vspace{-0.5em}
	\caption{Top-1 accuracy \emph{v.s.} FLOPs on ImageNet dataset.}
	\label{fig:speed}
	\vspace{-1em}
\end{figure}

\begin{figure}[htb]
	\centering
	\small
	\includegraphics[width=0.9\linewidth]{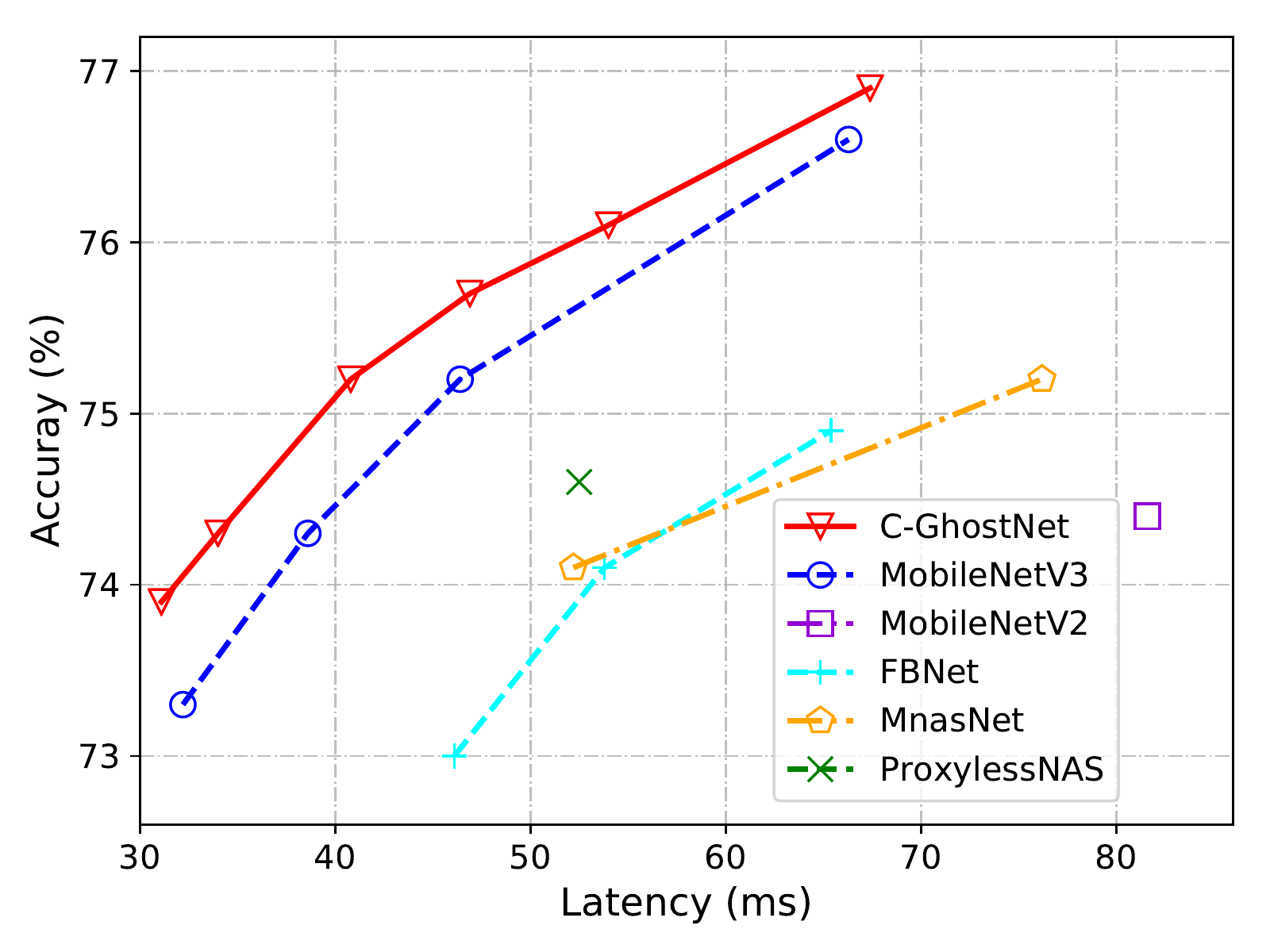}
	\vspace{-0.5em}
	\caption{Top-1 accuracy \emph{v.s.} latency on ImageNet dataset.}
	\label{fig:speed2}
	\vspace{-0.5em}
\end{figure}

\paragraph{Actual Inference Speed.} Since the proposed C-GhostNet is designed for mobile applications, we further measure the actual inference speed of C-GhostNet on an ARM-based mobile phone using the TFLite tool~\citep{tensorflow}. Following the common settings in ~\citep{mobilenet,mobilev2}, we use single-threaded mode with batch size 1. From the results in Figure~\ref{fig:speed2}, we can see that C-GhostNet obtain about 0.5\% higher top-1 accuracy than MobileNetV3 with the same latency, and C-GhostNet need less runtime to achieve similar performance. For example, C-GhostNet with 75.2\% accuracy only has 40 ms latency, while MobileNetV3 with similar accuracy requires about 46 ms to process one image. Overall, our models generally outperform the famous state-of-art models, \ie, MobileNet series~\citep{mobilenet,mobilev2,mobilenetv3}, ProxylessNAS~\citep{proxylessnas}, FBNet~\citep{fbnet}, and MnasNet~\citep{mnasnet}.

\paragraph{Ablation Study.}
From the results in Table~\ref{tab:C-GhostNet-imagenet}, C-GhostNet has more parameters than other networks with similar FLOPs. We evaluate other manners to include more parameters without introducing more FLOPs, \eg, using more SE modules, or using more channels and a smaller input resolution. We construct Net-A and Net-B which adopt the same network architecture as C-GhostNet in Table~\ref{tab:C-GhostNet} and replace C-Ghost bottlenecks with inverted residual bottlenecks. SE module is applied in every block of Net-A. The input resolution of Net-B is set as 192$\times$192. We adjust the width of Net-A and Net-B to obtain similar FLOPs to C-GhostNet. From the results in Table~\ref{tab:ghostnet-ablation}, the other manners to increase parameters perform weaker than our C-GhostNet, indicating the effectiveness of C-GhostNet.

C-Ghost module utilizes the depth-wise convolution to generate ghost features, which partly makes the model "deeper". Here we verify that the performance gain is not coming from simply making model deeper. We construct Net-C and Net-D with the same network architecture as C-GhostNet in Table~\ref{tab:C-GhostNet} and replace C-Ghost bottlenecks with inverted residual bottlenecks. We add 1 more and 2 more blocks for each stage in Net-C and Net-D respectively. To keep the FLOPs similar, we adjust the width of Net-C and Net-D. From the results in Table~\ref{tab:ghostnet-ablation}, the deeper models perform much worse than our C-GhostNet, that is to say, the accuracy gain of C-GhostNet is not from simply making model deeper.

\begin{table}[htb]
	\centering
	\small
	\renewcommand\arraystretch{1.0}
	\caption{Ablation study of C-GhostNet on ImageNet dataset.}
	\vspace{0.1em}
	\label{tab:ghostnet-ablation}
	\setlength{\tabcolsep}{7pt}{
		\begin{tabular}{l|c|c|c}
			\toprule[1.5pt]
			Model    & Params (M) &  \tabincell{c}{FLOPs (M)} & Top-1 (\%) \\
			\midrule
			C-GhostNet 0.5$\times$ & 2.6 & 42 & 66.2  \\
			Net-A & 2.6 & 42 & 64.2  \\
			Net-B & 2.6 & 46 & 65.7  \\
			Net-C & 2.1 & 44 & 62.3  \\
			Net-D & 2.0 & 42 & 61.0  \\
			\bottomrule[1pt]
		\end{tabular}
	}
	\vspace{-1em}
\end{table}

\paragraph{Generalization to Object Detection.} 
In order to further evaluate the generalization ability of C-GhostNet, we conduct object detection experiments on MS COCO dataset. We use the \emph{trainval35k} split as training data and report the results in mean Average Precision (mAP) on \emph{minival} split, following~\citep{fpn,retinanet}. The one-stage RetinaNet~\citep{retinanet} is used as our framework and C-GhostNet acts as a drop-in replacement for the backbone feature extractor. We train all the models using SGD for 12 epochs from ImageNet pretrained weights with the hyper-parameters suggested in~\citep{fpn,retinanet}. The input images are resized to a short side of 800 and a long side not exceeding 1333. Table~\ref{tab:voc2007} shows the detection results, where the FLOPs are calculated using $224\times224$ images as common practice and the latency is measured on a smartphone as that in the above section. With significantly lower computational costs, C-GhostNet achieves similar mAP with MobileNetV2 and MobileNetV3.

\begin{table}[htb]
	\centering
	\small
	\renewcommand\arraystretch{1.0}
	\caption{C-GhostNet Results on MS COCO dataset.}
	\vspace{0.1em}
	\label{tab:voc2007}
	\setlength{\tabcolsep}{5pt}{
		\begin{tabular}{l|c|c|c}
			\toprule[1.5pt]
			Backbone      &  \tabincell{c}{Backbone FLOPs} & ARM Latency & mAP \\
			\midrule
			MobileNetV2 1.0$\times$   &  300M  & 2.7 s & 26.7  \\
			MobileNetV3 1.0$\times$    & 219M  & 2.8 s & 26.4  \\
			C-GhostNet 1.1$\times$ & 164M & 2.5 s & 26.6  \\
			\bottomrule[1pt]
		\end{tabular}
	}
	\vspace{-1em}
\end{table}

\textbf{
	\subsection{GPU-Efficient G-Ghost Structure}
}
We apply the proposed G-Ghost stage on various networks, and conduct experiments on image recognition and object detection tasks to verify its effectiveness.

\subsubsection{CIFAR10 Experiments}
We first verify the effectiveness of our method and conduct ablation studies on CIFAR10 dataset. We use the widely-used ResNet56 as the baseline network for CIFAR10 classification.

\paragraph{Main Results.}
By upgrading all the stages with our G-Ghost stage, we construct G-Ghost-ResNet56 model. Here we set the ghost ratio $\lambda=0.5$ and the effect of $\lambda$ can be seen in the following paragraph. From the results in Table~\ref{tab:res56}, G-Ghost-ResNet56 without mix operation outperform the baseline ResNet56-0.7$\times$ by a significant margin. By adding the mix operation for intrinsic feature aggregation, the performance is further improved which is 0.78\% higher than baseline. G-Ghost-ResNet56 w/ mix reduces the FLOPs by 2.32$\times$ with only 0.21\% accuracy drop compared to the vanilla ResNet56.

\begin{table}[htp]
	\small 
	\centering
	\caption{Main results of G-Ghost stage on CIFAR10.}\label{tab:res56}
	\renewcommand{\arraystretch}{1.0}
	\setlength{\tabcolsep}{6pt}{
		\begin{tabular}{l|c|c|c}
			\toprule[1.5pt]
			Model & Params (M) & FLOPs (M) & Acc (\%) \\
			\midrule
			ResNet56  & 0.85 & 125.0 & 94.1 \\
			ResNet56-0.7$\times$  & 0.41 & 63.3 & 93.2 \\
			G-Ghost w/o mix  & 0.36 & 53.7 & 93.7 \\
			G-Ghost w/ mix  & 0.37 & 53.8 & 93.9 \\
			\bottomrule[1pt]
		\end{tabular}
	}
	\vspace{-1.0em}
\end{table}

\paragraph{Comparison with Other Methods.}
We also compare our method with other state-of-the-art model compression methods or efficient models, including GAL~\citep{gal}, FPGM~\citep{fpgm}, LEGR~\citep{legr}, HRank~\citep{hrank}. As shown in Table~\ref{tab:res56-sota}, our method achieves higher accuracy with fewer FLOPs and parameters than other methods.

\begin{table}[htp]
	\vspace{-1em}
	\centering
	\small 
	\caption{Comparison with other methods on CIFAR10.}\label{tab:res56-sota}
	\renewcommand{\arraystretch}{1.0}
	\setlength{\tabcolsep}{4pt}{
		\begin{tabular}{l|c|c|c}
			\toprule[1.5pt]
			Model & Params (M) & FLOPs (M) & Acc (\%) \\
			\midrule
			ResNet56  & 0.85 & 125.0 & 94.1 \\
			\midrule
			PFEC~\citep{l1-pruning} & 0.73 & 90.9 & 93.1 \\
			CP~\citep{cp} & - & 62.5 & 92.0 \\
			GAL~\citep{gal} & 0.75 & 78.3 & 93.0 \\
			FPGM~\citep{fpgm} & - & 59.4 & 93.5 \\
			LEGR~\citep{legr} & - & 58.9 & 93.7 \\
			HRank~\citep{hrank} & 0.49 & 62.7 & 93.2 \\
			G-Ghost w/ mix  & 0.37 & 53.8 & \textbf{93.9} \\
			\bottomrule[1pt]
		\end{tabular}
	}
	\vspace{-1.0em}
\end{table}

\paragraph{Importance of Cheap Operations.} The cross-layer cheap operation is one of the key-points in our method. We evaluate the importance of the cheap operation by 1) replacing it with identity mapping, 2) remove it and produce the removed ghost features from the last layer. From Table~\ref{tab:cheap}, we can see that whether using identity mapping or removing cheap operation performs worse than using 1$\times$1 convolution as cheap operation.

\begin{table}[htp]
	\vspace{-0em}
	\small 
	\centering
	\caption{Evaluating the importance of cheap operations.}\label{tab:cheap}
	\renewcommand{\arraystretch}{1.0}
	\setlength{\tabcolsep}{8pt}{
		\begin{tabular}{l|c|c|c}
			\toprule[1.5pt]
			Cheap Op & Params (M) & FLOPs (M) & Acc (\%) \\
			\midrule
			Identity  & 0.36 & 53.3 & 93.3 \\
			No  & 0.41 & 60.8 & 93.2 \\
			1$\times$1 Conv  & 0.36 & 53.7 & 93.7 \\
			\bottomrule[1pt]
		\end{tabular}
	}
	\vspace{-1.5em}
\end{table}

\paragraph{Type of Cheap Operations.} There are various candidates for the cheap operation, such as convolution with different kernel size. We test different types of cheap operations where the ghost ratio $\lambda$ is set as 0.5 and show the results in Table~\ref{tab:type}. The 3$\times$3 Conv obtains the best accuracy. Including more parameters as in 5$\times$5 Conv does not bring in higher accuracy. The 1$\times$1 Conv achieves a good performance with fewer FLOPs and parameters, so we adopt it as cheap operation in the following experiments unless otherwise specified.

\begin{table}[htp]
	\small 
	\centering
	\caption{Effect of different types of cheap operations.}\label{tab:type}
	\renewcommand{\arraystretch}{1.0}
	\setlength{\tabcolsep}{8pt}{
		\begin{tabular}{l|c|c|c}
			\toprule[1.5pt]
			Cheap Op & Params (M) & FLOPs (M) & Acc (\%) \\
			\midrule
			1$\times$1 Conv  & 0.36 & 53.7 & 93.67 \\
			3$\times$3 Conv  & 0.38 & 56.9 & 93.74 \\
			5$\times$5 Conv  & 0.43 & 63.2 & 93.63  \\
			\bottomrule[1pt]
		\end{tabular}
	}
	\vspace{-1em}
\end{table}

\begin{figure}[htp]
	\vspace{-0em}
	\centering
	\includegraphics[width=0.8\linewidth]{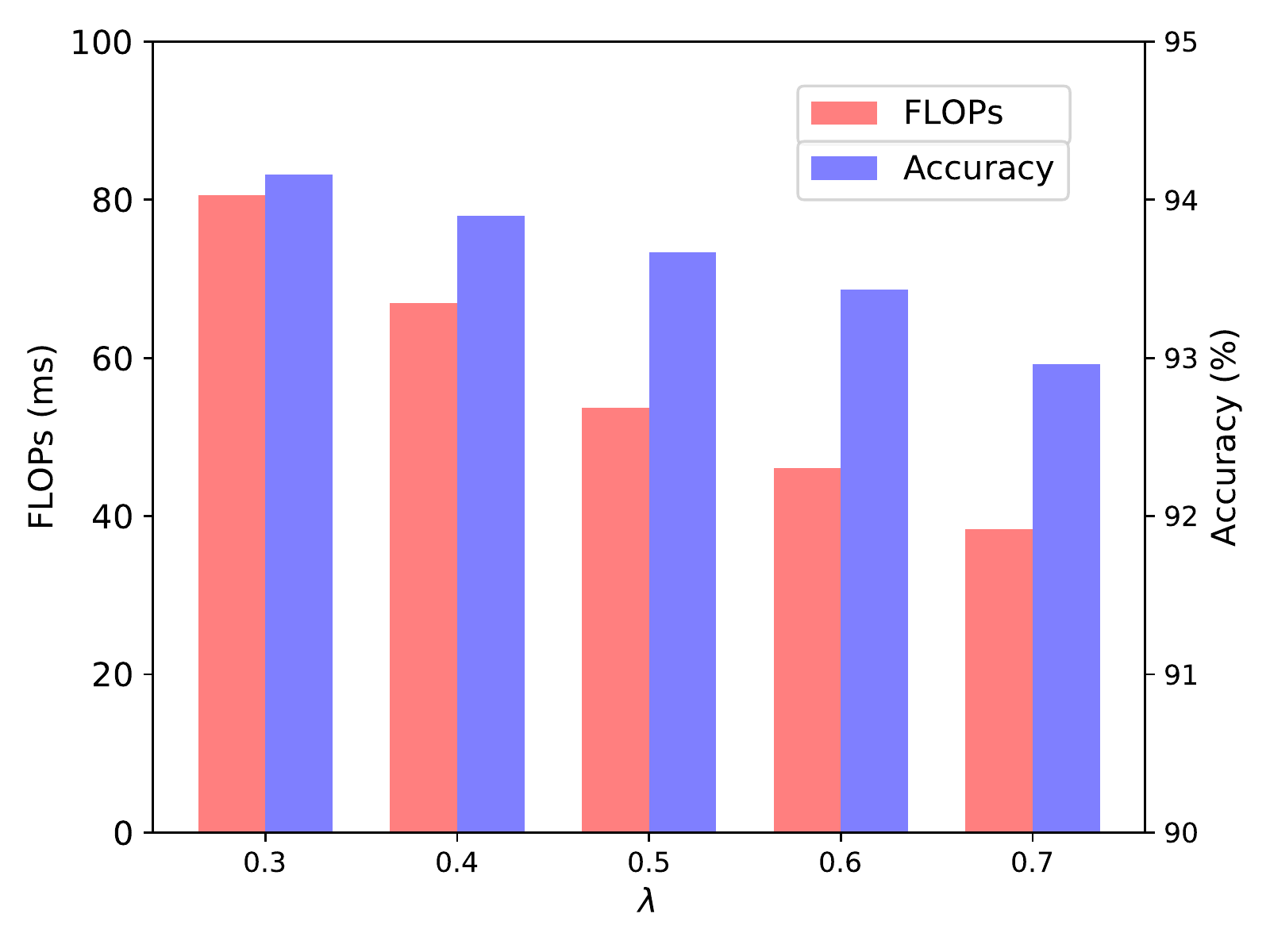}
	\vspace{-1em}
	\caption{The impact of ghost ratio $\lambda$ in G-Ghost stage.}
	\label{Fig:lambda}
	\vspace{-1em}
\end{figure}

\paragraph{Ratio of Cheap Operations.} The ghost ratio $\lambda$ is a hyper-parameter controlling the ratio of cheap features and directly influences the compression ratio as in Eq.~\ref{eq:ratio1} and Eq.~\ref{eq:ratio2}. We use 1$\times$1 convolution as the cheap operation and tune $\lambda$ in $[0.3,0.4,0.5,0.6,0.7]$. The results are shown in Figure~\ref{Fig:lambda}. As $\lambda$ increases, the FLOPs decrease largely while the accuracy drops gradually. We usually set the ghost ratio $\lambda$ as 0.5 due to the trade-off between computational cost and performance unless otherwise specified.

\subsubsection{ImageNet Experiments}
To verify the generalization of the proposed method, we test it on the large-scale ImageNet benchmark under settings of various model sizes.

\begin{table}[htp]
	\vspace{-0em}
	\small 
	\centering
	\caption{ResNet results on ImageNet. Latency is tested on an NVIDIA Tesla V100 GPU with a batch size of 64.}\label{tab:resnet}
	\scriptsize
	\renewcommand{\arraystretch}{1.0}
	\setlength{\tabcolsep}{6pt}{
		\begin{tabular}{l|c|c|c|c|c|c}
			\toprule[1.5pt]
			\multirow{2}*{Model} & Params & FLOPs & Acts & Latency & Top1 & Top5 \\
			& (M) & (B) & (M) & (ms) & (\%) & (\%) \\
			\midrule
			ResNet34  & 21.8 & 3.7 & 3.7 & 29.1  & 73.3 & 91.4 \\
			ResNet34-0.8$\times$  & 14.0 & 2.3 & 3.0 & 24.0  & 71.7 & 90.3 \\
			G-Ghost w/o mix  & 14.3 & 2.3 & 3.3 & 24.1  & 72.8 & 90.9 \\
			G-Ghost w/ mix  & 14.6 & 2.3 & 3.3 & 24.5  & 73.1 & 91.2 \\
			\bottomrule[1pt]
		\end{tabular}
	}
	\vspace{-1em}
\end{table}

\subsubsection{ResNet}
The proposed G-Ghost stage is GPU-efficient and can be used to accelerate the existing networks. We apply our method by replacing all the stage structure in ResNet34 using our G-Ghost stage. The results are shown in Table~\ref{tab:resnet}. Our G-Ghost-ResNet34 without mix operation outperforms ResNet34-0.8$\times$ by a significant margin with similar FLOPs and GPU latency. Including mix operation further improve the performance with negligible extra computational cost. Finally, our G-Ghost-ResNet34 with mix operation achieves comparable accuracy with original ResNet34 by reducing about 16\% GPU latency.

\paragraph{Comparison with Network Pruning.}
Network pruning is a kind of widely-used method for simplifying neural networks by cutting out the unimportant channels. We compare G-Ghost-ResNet34 with the state-of-the-art pruning methods including SFP~\citep{sfp}, PFEC~\citep{l1-pruning}, Taylor-FO-BN~\citep{molchanov2019importance} and FPGM~\citep{fpgm}. From the results in Table~\ref{tab:resnet-sota}, We can see that our G-Ghost-ResNet34 outperforms the other methods which shows the superiority of our method.

\begin{table}[htp]
	\vspace{-0em}
	\centering
	\small
	\caption{Comparison with other methods on ImageNet.}\label{tab:resnet-sota}
	\scriptsize 
	\renewcommand{\arraystretch}{1.0}
	\setlength{\tabcolsep}{3.5pt}{
		\begin{tabular}{l|c|c|c|c|c}
			\toprule[1.5pt]
			\multirow{2}*{Model}  & Params & FLOPs & Acts & Top1 & Top5 \\
			& (M) & (B) & (M) & (\%) \\
			\midrule
			ResNet34  & 21.8 & 3.7 & 3.7 & 73.3 & 91.4 \\
			\midrule
			SFP~\citep{sfp}  & 13.1 & 2.2 & - & 71.8 & 90.3 \\
			PFEC~\citep{l1-pruning}  & 19.3 & 2.8 & - & 72.2 & - \\
			Taylor-FO-BN~\citep{molchanov2019importance}  & 17.2 & 2.8 & - & 72.8 & - \\
			FPGM~\citep{fpgm}  & 13.1 & 2.2 & - & 72.6 & 91.1 \\
			ABCPruner~\citep{lin2020channel} & 10.1 & 2.2 & - & 71.0 & 90.1 \\
			G-Ghost w/ mix  & 14.6 & 2.3 & - & \textbf{73.1} & \textbf{91.2} \\
			\bottomrule[1pt]
		\end{tabular}
	}
	\vspace{-1.0em}
\end{table}

\begin{figure}[htp]
	\centering
	\includegraphics[width=1.0\linewidth]{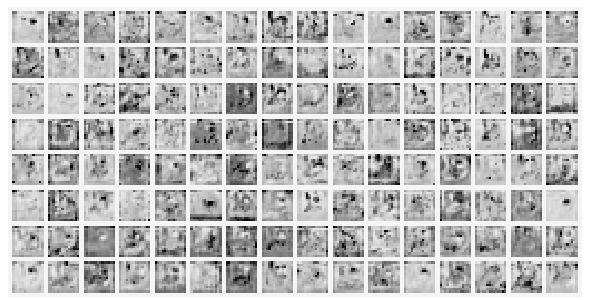}
	\vspace{-1em}
	\caption{The feature maps of the third stage of ResNet34.}
	\label{Fig:res34-layer2}
	\vspace{-1em}
\end{figure}


\begin{figure}[htp]
	\vspace{-0em}
	\centering
	\includegraphics[width=1.0\linewidth]{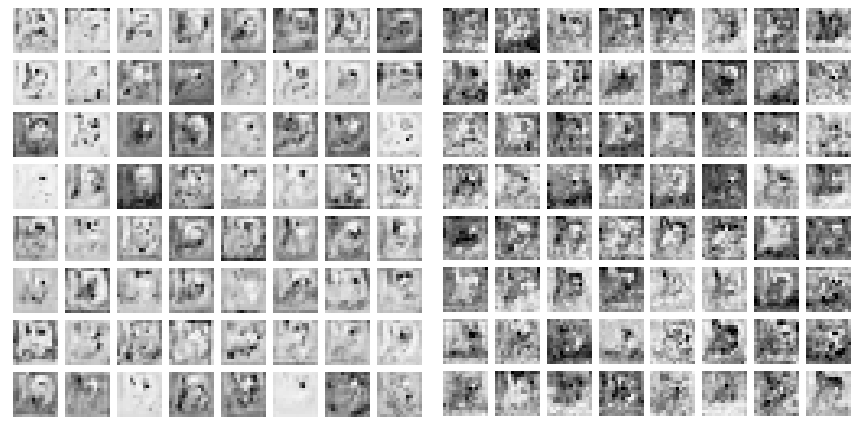}
	\vspace{-1em}
	\caption{The feature maps of the third stage of G-Ghost-ResNet34. Left: features from complicated path, right: features from cheap operation.}
	\label{Fig:mixghost-res34-maps}
	\vspace{-1.em}
\end{figure}


\paragraph{Visualization of Feature Maps.}
To better understand the effect of the proposed G-Ghost stage, we show the feature maps of the original ResNet and the G-Ghost-ResNet in Figure~\ref{Fig:res34-layer2} and Figure~\ref{Fig:mixghost-res34-maps}, respectively. The pattern in features of G-Ghost-ResNet34 (Figure~\ref{Fig:mixghost-res34-maps}) is also similar to that of the original ResNet34 (Figure~\ref{Fig:res34-layer2}) so the cheap operation achieve its goal to generate informative features. The reduced complexity does not harm the capacity of the network.

\subsubsection{RegNet}
RegNet is a series of networks derived by exploring the structure aspect of network design~\citep{regnet}, which is more efficient than ResNet. We test our G-Ghost stage by replacing all the stages with more than two blocks in RegNetX with our G-Ghost stage. We show the results in Table~\ref{tab:regnet}. From Table~\ref{tab:regnet}, we can see that our G-Ghost stage achieves 1.1\% higher top-1 accuracy than RegNetX-3.2GF-0.75$\times$ with similar inference latency. 

\begin{table}[htp]
	\vspace{-0em}
	\small 
	\centering
	\caption{RegNet Results on ImageNet. Latency is tested on an NVIDIA Tesla V100 GPU with batch size of 64.}\label{tab:regnet}
	\scriptsize
	\renewcommand{\arraystretch}{1.0}
	\setlength{\tabcolsep}{4.5pt}{
		\begin{tabular}{l|c|c|c|c|c|c}
			\toprule[1.5pt]
			\multirow{2}*{Model} & Params & FLOPs & Acts & Latency & Top1 & Top5 \\
			& (M) & (B) & (M) & (ms) & (\%) & (\%) \\
			\midrule
			RegNetX-3.2GF  & 15.3 & 3.2 & 11.4 & 59.1  & 78.2 & 94.0 \\
			RegNetX-3.2GF-0.75$\times$  & 8.8 & 1.8 & 8.6 & 46.1  & 76.3 & 93.0 \\
			G-Ghost w/o mix  & 9.7 & 1.8 & 8.7 & 43.9  & 76.8 & 93.2 \\
			G-Ghost w/ mix   & 10.5 & 1.8 & 8.7 & 44.4  & 77.8 & 93.7 \\
			\bottomrule[1pt]
		\end{tabular}
	}
	\vspace{-1.0em}
\end{table}

\begin{table*}[htp]
	\small 
	\centering
	\caption{GhostX-RegNet Results on ImageNet. Latency is tested on an NVIDIA Tesla V100 GPU with batch size of 64.}\label{tab:regnet-imagenet}
	\renewcommand{\arraystretch}{1.0}
	\setlength{\tabcolsep}{7pt}{
		\begin{tabular}{l|c|c|c|c|c|c}
			\toprule[1.5pt]
			Model & Params (M) & FLOPs (B) & Acts (M) & GPU latency (ms) & Top-1 (\%) & Top-5 (\%) \\
			\midrule
			ResNet-50~\citep{resnet}  & 25.6 & 4.1 & 11.1 & 54.7  & 76.2 & 92.9 \\
			ResNet-101~\citep{resnet}  & 44.6 & 7.8 & 16.2 & 93.5  & 77.4 & 93.6 \\
			ResNet-152~\citep{resnet}  & 60.2 & 11.5 & 22.6 & 135.0  & 78.3 & 94.1 \\
			\midrule
			Densenet-121~\citep{densenet}  & 8.0 & 2.8 & 6.9 & 57.3  & 74.6 & 92.2 \\
			Densenet-169~\citep{densenet}  & 14.2 & 3.4 & 7.3 & 73.1  & 76.0 & 93.0 \\
			Densenet-201~\citep{densenet}  & 20.0 & 4.3 & 7.9 & 95.4  & 77.2 & 93.6 \\
			Densenet-161~\citep{densenet}  & 28.7 & 7.7 & 11.1 & 130.6  & 77.6 & 93.8 \\
			\midrule	
			ResNeXt-50-32x4d~\citep{resnext}  & 25.0 & 4.0 & 14.4 & 75.6  & 77.8 & 93.7 \\
			ResNeXt-101-32x4d~\citep{resnext}  & 42.3 & 8.0 & 21.2 & 131.4  & 78.8 & 94.3 \\
			\midrule
			InceptionV3~\citep{inceptionv3}  & 26.0 & 5.7 & 4.6 & 68.5  & 77.4 & 93.6 \\	
			\midrule
			EfficientNet-B3~\citep{efficientnet}  & 12.0 & 1.8 & 23.8 & 88.4  & 77.5 & - \\
			EfficientNet-B4~\citep{efficientnet}  & 19.0 & 4.2 & 48.5 & 185.1  & 78.8 & - \\
			EfficientNet-B5~\citep{efficientnet}  & 30.0 & 9.9 &  98.9 & 380.6  & 78.5 & - \\
			\midrule
			RegNetX-1.6GF~\citep{regnet}  & 9.2 & 1.6 & 7.9 & 40.8  & 77.0 & - \\
			RegNetX-3.2GF~\citep{regnet}  & 15.3 & 3.2 & 11.4 & 59.1  & 78.3 & - \\
			RegNetX-4.0GF~\citep{regnet}  & 22.1 & 4.0 & 12.2 & 71.1  & 78.6 & - \\
			RegNetX-8.0GF~\citep{regnet}  & 39.6 & 8.0 & 14.1 & 92.2  & 79.3 & - \\
			RegNetX-16GF~\citep{regnet}  & 54.3 & 15.9 & 25.5 & 165.2  & 80.0 & - \\
			\midrule
			GhostX-RegNetX-3.2GF  & 10.5 & 1.8 & 8.7 & \textbf{44.4}  & \textbf{77.8} & 93.9 \\
			GhostX-RegNetX-4.0GF  & 15.7 & 2.2 & 9.5 & \textbf{52.1}  & \textbf{78.6} & 94.3 \\
			GhostX-RegNetX-8.0GF  & 24.6 & 4.2 & 10.4 & \textbf{62.9}  & \textbf{79.0} & 94.5 \\
			GhostX-RegNetX-16GF  & 32.4 & 9.0 & 19.5 & \textbf{110.9}  & \textbf{79.9} & 95.1 \\
			\bottomrule[1pt]
		\end{tabular}
	}
	\vspace{-1.0em}
\end{table*}

\begin{figure}[htp]
	\vspace{-0em}
	\centering
	\includegraphics[width=0.95\linewidth]{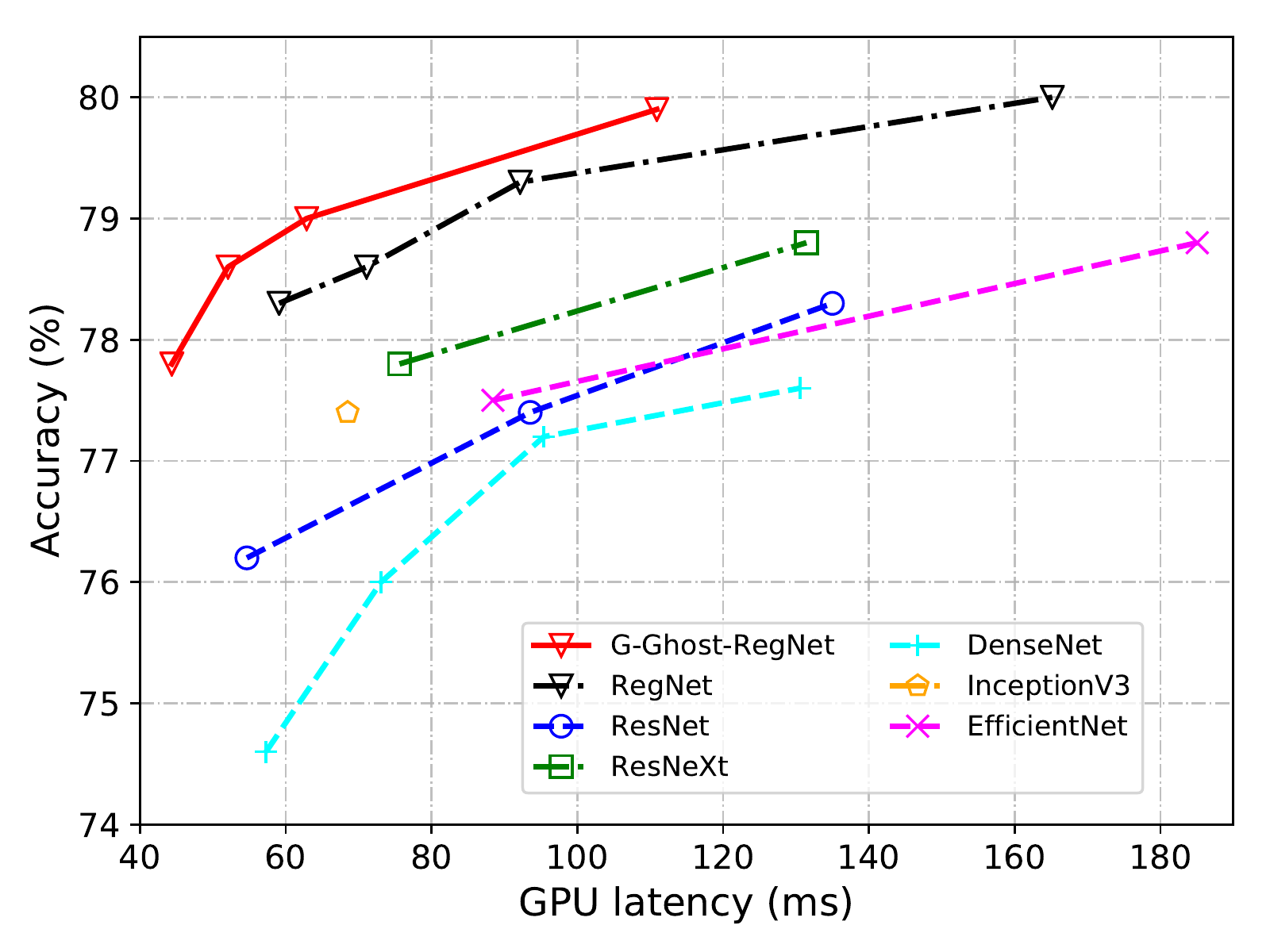}
	\vspace{-0.5em}
	\caption{Accuracy \emph{v.s.} GPU latency of G-Ghost-RegNet and other representative models.}
	\label{Fig:regnet-acc-gpu}
	\vspace{-1em}
\end{figure}

We also compare G-Ghost-RegNet with the other modern CNN structures, such as RegNet~\citep{regnet}, ResNet~\citep{resnet}, ResNeXt~\citep{resnext}, DenseNet~\citep{densenet}, InceptionV3~\citep{inceptionv3} and EfficientNet~\citep{efficientnet}. EfficientNet is trained under the same setting as in~\citep{regnet} for fair comparison. The detailed results are shown in Table~\ref{tab:regnet-imagenet}. Our G-Ghost-RegNet achieves the optimal accuracy-FLOPs trade-off and accuracy-latency trade-off (Figure~\ref{Fig:regnet-acc-gpu}). Note that although EfficientNet has a good trade-off between accuracy and FLOPs, it requires a large GPU latency, which is consistent with our analysis in the introduction.


\paragraph{Generalization to Object Detection.}
We choose the typical one-stage detector RetinaNet~\citep{retinanet} as basic framework, and use the proposed G-Ghost stage as a drop-in replacement for the backbone feature extractor. All the models are trained with 1$\times$ learning rate schedule (12 epochs). Table~\ref{tab:coco} shows the results on COCO minival. Our G-Ghost stage can increase the GPU speed from 21.9 FPS to 25.9 FPS with a small mAP drop. In another hand, G-Ghost-RegNetX-3.2GF suppresses ResNet50 and RegNetX-3.2GF-0.75$\times$ by a significant margin, and meanwhile achieves a faster inference speed, which demonstrates the effectiveness and generalization ability of our G-Ghost backbone.

\begin{table}[htp]
	\small 
	\centering
	\caption{Evaluation results based on RetinaNet on COCO minival. FLOPs denotes the FLOPs of backbone at 224$\times$224 input size. The inference speed is tested on an NVIDIA Tesla V100 GPU.}\label{tab:coco}
	\renewcommand{\arraystretch}{1.0}
	\setlength{\tabcolsep}{4pt}{
		\begin{tabular}{l|c|c|c|c|c|c}
			\toprule[1.5pt]
			Backbone & mAP & $\rm{AP}_{S}$ & $\rm{AP}_{M}$ & $\rm{AP}_{L}$ & FLOPs & FPS \\
			\midrule
			RegNetX-3.2GF & 39.1 & 22.6 & 43.5 & 50.8 & 3.2B & 21.9 \\
			\midrule
			ResNet50  & 36.8 & 20.5 & 40.6 & 48.5 & 4.1B & 22.4 \\
			ResNeXt50  & 37.6 & 21.1 & 41.4 & 49.4 & 4.2B & 19.6 \\
			RegNetX{\scriptsize-3.2GF-0.75$\times$} & 37.7 & 21.2 & 41.6 & 49.5 & 1.8B & 25.2 \\
			G-Ghost{\scriptsize-RegNetX-3.2GF}   & \textbf{38.4} & 22.3 & 42.3 & 50.1 & 1.8B & \textbf{25.9} \\
			\bottomrule[1pt]
		\end{tabular}
	}
	\vspace{-0.0em}
\end{table}

\subsubsection{Mobile Networks}
\begin{figure}[htp]
	\vspace{-0em}
	\centering
	\includegraphics[width=0.95\linewidth]{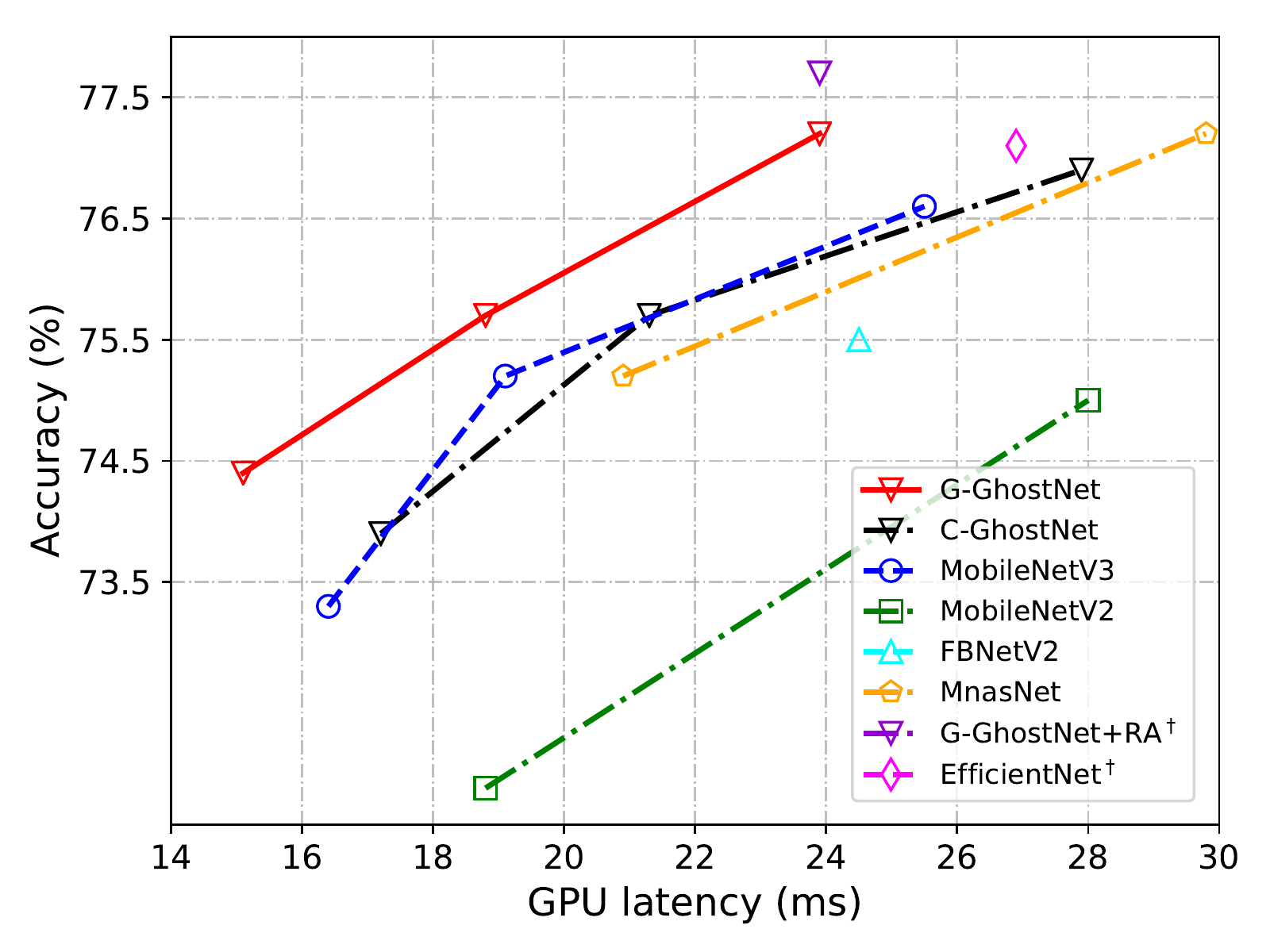}
	\vspace{-0.5em}
	\caption{Accuracy \emph{v.s.} GPU latency of G-GhostNet and other representative models. GPU Latency is tested on an NVIDIA Tesla V100 GPU with batch size of 64. $^{\dagger}$Advanced auto-augment is applied.}
	\label{Fig:gghostnet-acc-gpu}
	\vspace{-1em}
\end{figure}

Most of the existing lightweight models are designed with the aim to reduce FLOPs and latency on CPUs, but neglect the efficiency on GPUs. Our G-GhostNet remould the C-GhostNet for improve the GPU speed. We train G-GhostNet with the similar training setting as C-GhostNet and other mobile networks. The comparison of G-GhostNet and other state-of-the-art lightweight models including MobileNets~\citep{mobilev2,mobilenetv3}, FBNetV2~\citep{fbnetv2} and EfficientNet~\citep{efficientnet}, is shown in Figure~\ref{Fig:gghostnet-acc-gpu}. Although G-GhostNet has larger FLOPs and higher CPU latency, it achieves higher accuracy than the other models with much lower latency on GPUs. By adding RandAugment~\citep{randaugment} in our model, G-GhostNet outperforms EfficientNet-B0 which also uses advanced auto-augment strategy by 0.6\%, with a faster inference on the GPU.

\section{Conclusion}\label{Conclusion}
In this paper, we present GhostNet series to accelerate the inference of convolutional neural networks on heterogeneous devices including CPU and GPU. First, we present a novel C-Ghost module for building CPU-efficient neural architectures. The basic C-Ghost module splits the original convolutional layer into two parts and utilizes fewer filters to generate several intrinsic feature maps. Then, a certain number of cheap operations will be further applied for generating ghost feature maps efficiently.  The experiments conducted on benchmark models and datasets illustrate that the proposed method is a plug-and-play module for converting original models to compact ones while remaining the comparable performance. In addition, the C-GhostNet built using the proposed new module outperforms state-of-the-art portable neural architectures, in terms of efficiency and accuracy.

To improve the GPU efficiency of convolutional neural networks, we further introduce a new G-Ghost stage. In addition to reducing FLOPs, GPU-efficient CNNs should also concern about run-time memory and avoid trivial computation. G-Ghost stage investigates the feature redundancy cross layers in convolutional structures. The output features in a stage are split into two parts: intrinsic features and ghost features. The intrinsic features are produced with original stage with fewer channels while the ghost features are generated using GPU-efficient cheap operations from the intrinsic ones. The information inherited in intermediate intrinsic features is additively aggregated into the ghost features to enhance the representation capacity. The proposed G-Ghost stage can be used to simplify most of the mainstream CNNs. The experiments on various networks and datasets have demonstrated the effectiveness of G-Ghost stage on GPUs.

\begin{acknowledgements}
This work was supported by NSFC (62072449, 61872241, 61632003), Macao FDCT Grant (0018/2019/AKP). Chang Xu was supported by the Australian Research Council under Project DP210101859 and the University of Sydney SOAR Prize.
\end{acknowledgements}

%
%

{\small
\bibliographystyle{spbasic}      
\bibliography{ref}   
}

\end{document}